\documentclass{article}

\usepackage{arxiv}

\usepackage[utf8]{inputenc} % allow utf-8 input
\usepackage[T1]{fontenc}    % use 8-bit T1 fonts
\usepackage{hyperref}       % hyperlinks
\usepackage{url}            % simple URL typesetting
\usepackage{booktabs}       % professional-quality tables
\usepackage{amsfonts}       % blackboard math symbols
\usepackage{nicefrac}       % compact symbols for 1/2, etc.
\usepackage{microtype}      % microtypography
\usepackage{graphicx}
\usepackage{natbib}
\usepackage{doi}

\usepackage{amsmath,amsfonts,bm}

\def\eqref#1{equation~\ref{#1}}
\def\1{\bm{1}}

\DeclareMathAlphabet{\mathsfit}{\encodingdefault}{\sfdefault}{m}{sl}
\SetMathAlphabet{\mathsfit}{bold}{\encodingdefault}{\sfdefault}{bx}{n}

\usepackage{tcolorbox}
\usepackage{longtable}
\usepackage{multirow}
\usepackage{todonotes}
\usepackage{tikz}
\usepackage{wrapfig}
\usetikzlibrary{positioning,shapes,arrows.meta}

\usepackage[font=small]{caption}
\usepackage{caption}
\usepackage{enumitem}
\setlist[enumerate,itemize]{leftmargin=2em, itemsep=0pt, parsep=0pt, topsep=0pt}

\usepackage[table]{xcolor}
\usepackage{array,tabularx}
\newcolumntype{Y}{>{\raggedright\arraybackslash}X} % normal ragged right
\newcolumntype{Z}{>{\raggedright\arraybackslash}X} % wide column for features

\newcolumntype{P}[1]{>{\raggedright\arraybackslash\hsize=#1\hsize}X}

\definecolor{colHeader}{HTML}{F5F5F5}
\definecolor{colO1}{HTML}{E8F5E9}   % light green
\definecolor{colO2}{HTML}{E3F2FD}   % light blue
\definecolor{colO3}{HTML}{FFF3E0}   % light orange

\newtheorem{theorem}{Theorem}[section]  % Numbered by section
\newtheorem{definition}[theorem]{Definition}

\usepackage{times}
\usepackage{latexsym}
\usepackage{fancyvrb}
\usepackage{fvextra}
\fvset{breaklines=true}
\usepackage{tcolorbox} % For colored boxes
\tcbuselibrary{listings,breakable}
\usepackage{lipsum}
\graphicspath{{imgs/}}

\usepackage{titlesec}
\titlespacing*{\paragraph}{0pt}{0.5ex plus 0.2ex minus 0.2ex}{0.5ex plus 0.1ex}

\usepackage{cleveref}

\title{Do LLM Agents Know How to Ground, Recover, and Assess? A Benchmark for Epistemic Competence in Information-Seeking Agents}

\author{Jiaqi Shao$^1$, Yuxiang Lin$^1$, Munish Prasad Lohani$^1$, Yufeng Miao$^2$, Bing Luo$^1$ \\
$^1$Duke Kunshan University\\
$^2$Microsoft AI}

\date{}

\renewcommand{\shorttitle}{SeekBench: Epistemic Competence Benchmark}

\usepackage{xspace}
\newcommand{\algo}{\textbf{SeekBench}\xspace}

\hypersetup{
pdftitle={Do LLM Agents Know How to Ground, Recover, and Assess? A Benchmark for Epistemic Competence in Information-Seeking Agents},
pdfsubject={cs.AI, cs.CL, cs.LG},
pdfauthor={Anonymous Authors},
pdfkeywords={Large Language Models, Search Agents, Epistemic Competence, Benchmark, Reasoning Quality, Evidence Grounding},
}

\begin{document}
\maketitle

\begin{abstract}
    Recent work has explored training Large Language Model (LLM) search agents with reinforcement learning (RL) for open-domain question answering (QA). However, most evaluations focus solely on final answer accuracy, overlooking how these agents reason with and act on external evidence.
    We introduce \algo, the first benchmark for evaluating the \textit{epistemic competence} of LLM search agents through step-level analysis of their response traces. \algo comprises 190 expert-annotated traces with over 1,800 response steps generated by LLM search agents, each enriched with evidence annotations for granular analysis of whether agents (1) generate reasoning steps grounded in observed evidence, (2) adaptively reformulate searches to recover from low-quality results, and (3) have proper calibration to correctly assess whether the current evidence is sufficient for providing an answer.
    Our analysis of state-of-the-art LLM search agents reveals critical behavioral gaps overlooked by traditional metrics, including specialized skills like Search-R1's synthesis capabilities. These findings expose distinct epistemic competencies that accuracy-only evaluations fail to capture, providing guidance for developing more capable and reliable agents.
\end{abstract}

% keywords can be removed

\section{Introduction}
Recent advances in Large Language Models (LLMs) have spurred a shift from models that require explicit prompting, such as chain-of-thought \citep{wei2022chain,yao2023react}, toward autonomous LLM agents~\citep{zhang2025landscape}. These agents learn to solve complex tasks by optimizing a reasoning policy with reinforcement learning (RL), enabling them to implicitly learn decision-making strategies without requiring step-by-step external guidance \citep{jaech2024openai,guo2025deepseek}.
Among these, \emph{search agents} are developed to tackle information-seeking problems by alternating between reasoning, using external search tools, and integrating evidence~\citep{xi2025survey}. 
The agent's responses produce multi-turn traces, which can be represented as:\looseness=-1
\begin{align} \mathcal{T} = \langle \tau_1, \tau_2, \ldots, \tau_T \rangle, \label{eq:trace} \end{align} 
where each turn $\tau_t$ consists of a tuple that can contain four possible steps: reasoning $r_t$, search $s_t$, evidence $e_t$, and answer $a_t$.
Non-final turns ($t < T$) are of the form $\tau_t = \langle r_t, s_t, e_t \rangle$, while the final turn ($t = T$) concludes with an answer, \emph{i}.\emph{e}., $\tau_T = \langle r_T, a_T \rangle$.

While the formal trace structure $\mathcal{T}$ captures the multi-turn execution of an agent's reasoning, search, and evidence integration, it does not address \emph{epistemic competence}—-the ability to reliably acquire, evaluate, and act upon knowledge in a justified manner~\citep{greene2016introduction,coeckelbergh2023democracy}.
Additionally, the current evaluation protocols for search 
agents rely predominantly on final-answer metrics such as 
exact match and F1 score \citep{searchr1, deepresearcher, 
websailor}.
However, in practice, \emph{agents may achieve high benchmark scores while exhibiting poor epistemic behaviors}, such as hallucinating unsupported claims, failing to recognize knowledge gaps, or lacking systematic approaches to information gathering and evaluation.

\begin{figure}[!t]
    \centering
    \begin{minipage}[t]{\textwidth}
    \centering
    \captionof{table}{Epistemic competencies and associated metrics.
    Each competency is quantified by a specific metric calculated from annotated features within the agent's trace (shown in the rightmost column), enabling systematic evaluation of reasoning quality, recovery behavior, and evidence-aligned decision-making.}
        \label{tab:competencies}
        \scriptsize
        \setlength{\tabcolsep}{3pt}
        \begin{tabular}{p{2.2cm}p{6.8cm}p{4.5cm}}
        \toprule
        \rowcolor{colHeader}
        \textbf{Competency (Type)} &
        \textbf{Definition \& Metric} &
        \textbf{Annotated feature(s)} \\
        \midrule
        \rowcolor{colO1}
        \textbf{Groundedness}\newline(Reasoning) &
        \begin{tabular}[c]{@{}p{6.8cm}@{}}Generate reasoning steps directly supported by retrieved information.\newline
        \textbf{Metric:} \emph{Reasoning Quality Index} (RQI, \Cref{sec:o1_groundness})\looseness=-1\end{tabular} &
        \begin{tabular}[c]{@{}p{4.5cm}@{}}\texttt{InformationSynthesis} /\newline \texttt{PlanFormation} / \texttt{StateAssessment} /\newline\texttt{grounding}\end{tabular} \\
        
        \rowcolor{colO2}
        \textbf{Recovery}\newline(Search) &
        \begin{tabular}[c]{@{}p{6.8cm}@{}}Adaptively reformulate queries when initial search results are insufficient.\newline
        \textbf{Metric:} \emph{Evidence Recovery Function} (ERF, \Cref{sec:o2_recovery})\looseness=-1\end{tabular} &
        \begin{tabular}[c]{@{}p{4.5cm}@{}}\texttt{Initial} / \texttt{Repeat} /\newline\texttt{FollowUp} / \texttt{Refined}\end{tabular}\\
        
        \rowcolor{colO3}
        \textbf{Calibration}\newline(Answer) &
        \begin{tabular}[c]{@{}p{6.8cm}@{}}Accurately assess whether the currently retrieved information is sufficient to answer the question.\newline
        \textbf{Metric:} \emph{Calibration Error} (CE, \Cref{sec:o3_calibration})\looseness=-1\end{tabular} &
        \begin{tabular}[c]{@{}p{4.5cm}@{}}\texttt{correct}\end{tabular} \\
        \bottomrule
        \end{tabular}
    \end{minipage}
    
    \vspace{0em} %
    
    \begin{minipage}[t]{\textwidth}
    \centering
    \includegraphics[width=\linewidth]{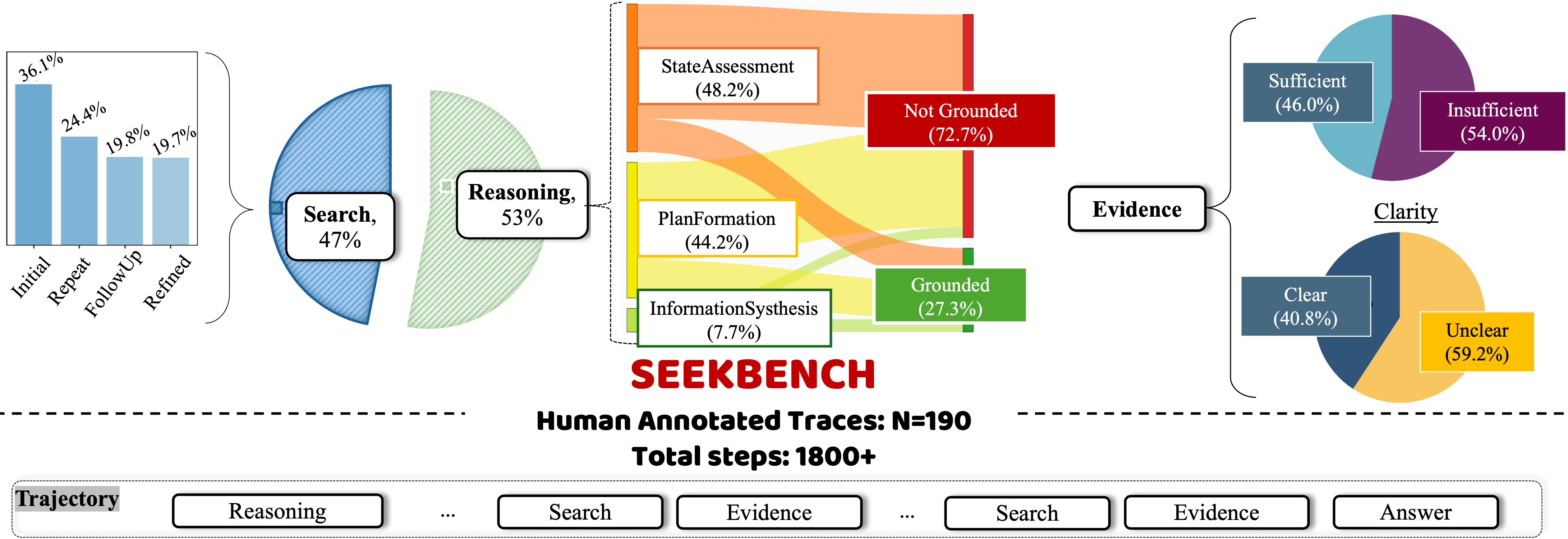}
    \captionof{figure}{\small Overview of the \algo dataset and annotation schema.
    Each trace comprises multi-turn steps annotated for process-level evaluation. We categorize agent behaviors into three main types: (1) \textbf{Search steps} that retrieve information, (2) \textbf{Reasoning steps} that process evidence and guide the investigation, and (3) \textbf{Evidence steps} that capture the quality and clarity of retrieved information. This structured annotation approach enables systematic measurement of how well agents handle information throughout their reasoning process.}
    \label{fig:dataset_overview}
    \end{minipage}
\end{figure}

Furthermore, as agent systems increasingly take on autonomous information-seeking and decision-making roles, focusing solely on final answers can be misleading because \textit{it fails to capture how agents navigate the information landscape and make decisions throughout the process}.
Without process-level evaluation, the impact of algorithmic innovations cannot be reliably assessed. These innovations may produce spurious performance gains, \textit{e}.\textit{g}., improvements in accuracy that come from rigid search steps or better answer synthesis, rather than from genuine advances in reasoning capabilities or information-seeking strategies.

To address answer-only evaluation limitations, we introduce \algo, a benchmark assessing search agents' epistemic behaviors that reflect sound evidence practices, including: (1) reasoning grounded in evidence, (2) adaptive search when information is insufficient, and (3) calibrated decision-making under uncertainty. Our framework centers on the concept of \textit{evidence 
state}, which captures information quality throughout the reasoning process. \algo includes 190 expert-annotated traces with over 1,800 steps, as shown in~\Cref{fig:dataset_overview}.

Our framework proceeds in three stages. First, we develop a robust, extensible annotation schema that captures both the functional role (reasoning, search, answer) and epistemic quality (e.g., evidence groundness and sufficiency) of each step. Second, we formalize three epistemic competencies that characterize: (1) \emph{evidence-grounded reasoning}, (2) \emph{adaptive evidence recovery}, and (3) \emph{evidence-aligned calibration}. 
Third, we design precise, interpretable metrics to quantify these competencies across diverse traces, as detailed in~\Cref{tab:competencies}.

\paragraph{Contributions.}
Our contributions are summarized as follows:
\begin{enumerate}
    \item \textbf{\algo: A Process-Level Benchmark.}
We introduce \algo, a benchmark for 
evaluating LLM search agents at the process level, 
providing a structured trace schema that decomposes agent 
traces into reasoning, search, evidence, and answer 
steps. Our annotated dataset achieves high human agreement (Cohen's Kappa $\kappa > 0.8$) and enables scalable evaluation through LLM judges with strong alignment to expert annotations ($\kappa > 0.7$).
    \item \textbf{Operational framework and metrics.} We formalize an \emph{evidence state} and three core \emph{epistemic competencies}—\emph{evidence-grounded reasoning}, \emph{evidence recovery}, and \emph{calibrated answering}—as measurable properties over agent traces.
    \item \textbf{Large-scale evaluation and findings.} Our evaluation across seven QA benchmarks (28,493 traces) reveals that RL agents excel at evidence gathering but struggle with reasoning. Standard accuracy metrics fail to reveal specific strengths between agents (e.g., Search-R1's synthesis vs. Base model's reasoning), which can be combined to enhance performance.
\end{enumerate}

\section{Related Work}

\paragraph{Reasoning Quality Assessment and Epistemic Competence}.
Recent work evaluates reasoning quality through various approaches: causal analysis of question-reasoning-answer triples \citep{paul2024making}, matching against golden reasoning chains \citep{meqa}, and graph-based methods that link step dependencies to answer correctness \citep{xiong2025mapping,nguyen2024direct,mukherjee2025premiseaugmented}. However, these methods focus on structural alignment rather than evaluating epistemic competence—the ability to reason about knowledge, uncertainty, and evidence. For AI systems that interact with external information sources, epistemic competence becomes critical for avoiding overconfidence, hallucination, and poor decision-making under uncertainty \citep{zhang2023siren}. However, existing frameworks fail to assess whether agents can ground reasoning in retrieved evidence or adapt search strategies when information is insufficient.

\paragraph{Search Agent Evaluation}.
Existing search agent evaluations primarily focus on final-answer metrics (exact match, F1, LLM-as-Judge) \citep{autocoa, r1searcher, deepresearcher, searchr1}, neglecting the epistemic processes underlying those answers. While some approaches examine intermediate steps—such as WebSailor's \citep{websailor} pass@k comparisons, AutoRefine's \citep{autorefine} ground-truth tracking, and InfoDeepSeek's \citep{infodeepseek} separation of retrieval from synthesis, they fail to assess critical epistemic competencies. 
As search agents increasingly mediate access to information and shape knowledge acquisition, this fundamental limitation in evaluation methodology creates significant blind spots in our understanding of agent capabilities. Current methods cannot evaluate whether agents: (1) ground reasoning in retrieved evidence, (2) adapt to poor search results, or (3) calibrate confidence based on evidence sufficiency—all essential for reliable information-seeking systems that can be trusted in real-world applications.

\paragraph{Process-Level Analysis Frameworks}.
Existing frameworks analyze reasoning processes through trace aggregation \citep{Ott2023}, step-level annotations \citep{wu2025kgtraces}, and reasoning taxonomies \citep{autorefine,satori25}. However, they lack formal definitions of epistemic competencies and quantitative metrics for measuring how agents handle uncertainty, adapt to insufficient information, and make evidence-aligned decisions. Our work addresses this gap by formalizing three core epistemic competencies (groundedness, recovery, and calibration) with precise mathematical definitions and large-scale evaluation protocols.

\section{Methodology}
To understand the information-seeking process of search agents, our framework connects observable behaviors with underlying competencies, which are then evaluated using quantitative metrics. First, we construct and validate an \textbf{annotation schema} that reliably labels observable behaviors in agent response traces (\Cref{sec:phase1}). Second, we analyze patterns in these annotations to identify three \textbf{fundamental epistemic competencies} (Table \ref{tab:competencies}) that are not directly observable but underlie the agent's epistemic performance--the degree to which the agent can effectively retrieve information and use it for justified decisions (\Cref{sec:phase2}). Finally, we translate these competencies into concrete \textbf{quantitative metrics} derived from the annotated features, enabling systematic measurement of these underlying epistemic competencies at scale (\Cref{sec:estimand_definitions}). Our approach draws on established qualitative research principles from \textit{Content Analysis} \citep{krippendorff2018content}, a systematic methodology for categorizing and interpreting patterns in data through rigorous coding procedures.

\subsection{Phase 1: Observable Features and Schema Construction}\label{sec:phase1}

The foundation of our methodology is a robust annotation schema for systematically labeling \emph{observable features} in agent response traces. Our development of \algo's schema follows an iterative, data-driven approach, grounded in established qualitative research principles from \emph{Content Analysis} \citep{krippendorff2018content}. 

During the initial exploratory phase, we closely examined a variety of agent traces and documented the key behaviors we observed. 
We noted that even within the same type of steps, search agents can serve distinct \textbf{functions}, that is, specific cognitive or operational role that a step plays within the agent's information-seeking process. 
For example, among the reasoning steps, some identified information gaps, while others summarized retrieved findings or formulated plans for future searches. Similarly, search steps might function as initial exploration, targeted verification, or follow-up investigation.
Alongside these functions, we identified critical \textbf{failure patterns} in 
agents' traces, such as reasoning without supporting evidence or executing repetitive search queries that failed to adapt.

From these observations, we developed an annotation schema for agent response steps with two key aspects:
(1) \textbf{Functional type} categorizing the step's cognitive purpose, e.g., for reasoning steps, this includes \texttt{InformationSynthesis}
(evidence integration), \texttt{PlanFormation} (search strategy 
development), and \texttt{StateAssessment} (knowledge gap 
identification).
(2) \textbf{Quality attribute} evaluating epistemic soundness, such as whether reasoning is \texttt{grounded} in evidence.
This structure captures both what the agent was 
doing and how well it was doing it.\looseness=-1

Following established \textit{Content Analysis} methodology \citep{krippendorff2018content}, we rigorously enhanced schema robustness through \emph{iterative refinement}. Three expert annotators independently coded 190 agent traces across three rounds of annotation, with inter-annotator reliability measured using Cohen's Kappa ($\kappa$) \citep{cohen1960coefficient}. For features exhibiting low agreement ($\kappa<0.5$), we either pruned features (when infrequent or ambiguous) or merged them (when conceptually overlapping). After this process, we reduced our initial 12 candidate annotation fields to 8 well-defined features with high interpretability and consistency across annotators.\looseness=-1

We further validated schema robustness using GPT-5 \citep{openai2025gpt5} to generate adversarial edge cases that expose boundary conditions (e.g., reasoning steps containing both factual claims and planning elements). This ensures the mutual exclusivity of our annotation definitions. Our final schema is detailed in~\Cref{fig:dataset_overview}.

Finally, we evaluated the schema on reasoning traces using both human experts and state-of-the-art LLM judges (GPT-4.1, GPT-4.1-mini \citep{openai2025gpt4_1}, and GPT-5 \citep{openai2025gpt5}) with standardized prompts that provide clear annotation guidelines and consistent evaluation criteria (see~Appendix~\ref{app:annotation_schema}).
The results demonstrate substantial agreement with human annotations (overall $\kappa=0.811$), confirming the schema's interpretability and consistency. LLM judges achieved strong alignment with human experts: GPT-4.1 ($\kappa=0.693$), GPT-4.1-mini ($\kappa=0.731$), and GPT-5 ($\kappa=0.754$). This substantial agreement across multiple LLM models and human annotators confirms 
the schema's clarity and establishes a reliable foundation for large-scale evaluation through LLM judges 
(see~Appendix~\ref{app:annotation_agreement} for details).\looseness=-1

\subsection{Phase 2: Latent Constructs and Competency Definition}\label{sec:phase2}

Our annotation analysis in Phase 1 revealed three distinct behavioral patterns in search agents: (1) variation in reasoning quality, with successful agents producing evidence-supported \texttt{grounded} reasoning while unsuccessful agents generated unsupported claims; (2) divergent strategies when facing poor search results, where effective agents adapted their search approach while ineffective agents persisted with repetitive queries; and (3) differences in decision timing, where some agents responded prematurely with insufficient evidence while others appropriately withheld answers until sufficient evidence was gathered.
To interpret these systematic behavioral differences, we applied the framework of \emph{latent construct inference} \citep{cronbach1955construct} to identify the underlying cognitive competencies for these observed patterns.

From these behavioral patterns, we derived three core epistemic competencies (Table \ref{tab:competencies}): \textbf{groundedness} measures alignment between reasoning steps and retrieved evidence; \textbf{recovery} evaluates an agent's ability to adapt search strategies after initial failures; and \textbf{calibration} assesses whether answering behavior appropriately corresponds to evidence quality.
These three core competencies constitute \textbf{epistemic competence}—the essential capability that enables search agents to reliably interact with external information sources. 
By systematically evaluating how agents \textit{seek}, \textit{reason with}, and \textit{make decisions} based on retrieved evidence, our framework provides a \emph{comprehensive assessment} of search agents' epistemic capabilities beyond traditional accuracy-based metrics.

\subsection{Phase 3: Competency Metrics and Operationalization}
\label{sec:estimand_definitions}
Following the concept of \textit{construct validity} \citep{cronbach1955construct} originally proposed in psychology, unobservable attributes (competencies) must be assessed through observable indicators (metrics) with demonstrated reliability and validity. 
In this section, we translated the three epistemic competencies in Phase 2 (defined in Table \ref{tab:competencies}) into quantitative metrics. 
The validity of our metrics is twofold:
(1) high inter-annotator agreement on the coded features (Phase 1), and (2) the correlation between evidence state (defined below) and answer accuracy (Section~\ref{sec:exp_calibration}).

We begin by formally defining \textbf{evidence state} in~Definition~\ref{def:evidence_state}, which encodes the sufficiency and clarity of evidence retrieved at a turn. This provides the foundation for evaluating all three epistemic competencies: (i) \textbf{groundedness} is assessed by determining whether reasoning is supported by evidence (\Cref{sec:o1_groundness}); (ii) \textbf{recovery} is measured by tracking improvements in evidence quality through search (\Cref{sec:o2_recovery}); and (iii) \textbf{calibration} is evaluated on whether the agent answers if and only if the evidence state is good (\Cref{sec:o3_calibration}).

\begin{definition}[Evidence State]\label{def:evidence_state}
Let $C_{i,t}, Q_{i,t} \in \{0,1\}$ denote the annotated \texttt{clarity} and \texttt{quality} of the retrieved evidence at turn $t$ of trace $i$, where: $C_{i,t} = 1$ if the evidence is clear (unambiguous and interpretable), and $Q_{i,t} = 1$ if the evidence is sufficient (contains enough information to address the query).
The \textbf{evidence state} $E_{i,t} \in \{0,1,2\}$ is defined as:
\begin{equation}
  E_{i, t} := C_{i, t} + Q_{i, t}, 
\end{equation}

$E_{i,t} = 0$ denotes \textbf{poor} evidence (unclear and insufficient), $E_{i,t} = 1$ denotes \textbf{partial} evidence (either clear or sufficient), and $E_{i,t} = 2$ denotes \textbf{good} evidence (both clear and sufficient).
\end{definition}

\subsubsection{Groundedness}\label{sec:o1_groundness}

To evaluate whether an agent's reasoning is verifiably supported by retrieved evidence, we assess the \textbf{groundedness} of each reasoning step via the \texttt{grounding} label. For each reasoning step at turn $t$ in trace $i$, the binary \texttt{grounding} label 
$G_{i,t} \in \{0,1\}$ indicates whether its factual content is supported by retrieved evidence.

To investigate the impact of the functional types of the reasoning steps, each reasoning step is also assigned a \texttt{type} $\mathcal{C}_{i,t} \in \{\texttt{IS},\texttt{PF},\texttt{SA}\}$, corresponding to \texttt{InformationSynthesis}, \texttt{PlanFormation}, or \texttt{StateAssessment}.

We formalize two metrics: the \textbf{model-level reasoning quality index (RQI)} (Definition~\ref{def:rqi_model}) and the \textbf{type-level RQI} (Definition~\ref{def:rqi_type}), both of which quantify groundedness by aggregating $G_{i,t}$ values and can be decomposed by the evidence state $E_{i,t}$.

\begin{definition}[Model-level Reasoning Quality Index (RQI)]\label{def:rqi_model}
Consider a fixed model evaluated on $N$ traces with index set $\mathcal{I} := \{1, \dots, N\}$. 
For each trace $i$, let $S_i = \{1, \ldots, T_i\}$ be the index set of reasoning steps.
Then, the model-level RQI is the average of trace-level groundedness scores:
\begin{equation}
\textstyle
\mathrm{RQI}_{\mathrm{model}} := \mathbb{E}_{i \in \mathcal{I}} [\mathrm{RQI}_i].
\end{equation}
where \( \mathrm{RQI}_i = \mathbb{E}_{t \in S_i}[G_{i,t}]\,,\)

\end{definition}

To better understand how reasoning quality depends on the strength of retrieved evidence, we decompose the RQI with evidence state $E_{i,t}$:
\begin{equation}
\mathrm{RQI}_i 
= \sum_{k=0}^{2} 
\underbrace{\mathbb{P}_{t \in S_i}(E_{i,t}=k)}_{\text{proportion of turns with evidence state }k}
\times 
\underbrace{\mathbb{E}_{t \in S_i}[G_{i,t} \mid E_{i,t}=k]}_{\text{expected groundedness given }E_{i,t}=k}\,
\end{equation}

Similarly, we can define the type-level RQI with reasoning type $c \in \{\texttt{IS}, \texttt{PF}, \texttt{SA}\}$:
\begin{definition}[Type-Level Reasoning Quality Index]\label{def:rqi_type}
For each trace $i$ and reasoning type $c \in {\texttt{IS}, \texttt{PF}, \texttt{SA}}$, let $S_i^{(c)} := \{ t \in S_i : \mathcal{C}_{i,t} = c \}$ denote the index set of steps of type $c$. The type-level RQI is the average of groundedness on type $c$:\looseness=-1
\begin{equation}\label{eq:rqi_type}
\mathrm{RQI}_{\mathrm{type}}^{(c)} := \mathbb{E}_{t \in S_i^{(c)}}\left[\mathrm{RQI}_i^{(c)}\right],
\end{equation}
where $\mathrm{RQI}_i^{(c)} := \mathbb{E}_{t \in S_i^{(c)}}[G_{i,t}]$.
This quantity admits an evidence-state decomposition analogous to the trace-level decomposition:
\begin{equation}\label{eq:rqi_type_evidence_decomposition}
\textstyle
\mathrm{RQI}_i^{(c)} 
= \sum_{k=0}^{2} \underbrace{\mathbb{P}_{t \in S_i^{(c)}}(E_{i,t}=k)}_{\text{prop. of reasoning type }c\text{ with evidence level }k}
\times \underbrace{\mathbb{E}_{t \in S_i^{(c)}}[G_{i,t} \mid E_{i,t}=k]}_{\text{expected type }c\text{ groundedness given }E_{i,t}=k}.
\end{equation}
\end{definition}

\subsubsection{Recovery}\label{sec:o2_recovery}

A fundamental challenge for LLM-based agents is recovering from information gaps or knowledge limitations, where initial queries yield insufficient information. Thus, an agent achieves high \textbf{recovery} when it utilizes adaptive search strategies to escape such states of poor evidence.

To capture this behavior, we use the evidence state $E_{i,t} \in \{0,1,2\}$ to track the sufficiency and clarity of retrieved information at each turn $t$ in trace $i$. We define a \emph{recovery event} (\Cref{eq:recovery_event}) as the first turn where the agent either (i) enters a high-evidence state ($E_{i,t} = 2$), or (ii) produces a correct answer. Formally:
\begin{equation}\label{eq:recovery_event}
    T_{\mathrm{recover},i} := \min \left\{t \in [1, T_i] : E_{i,t} = 2 \;\; \text{or} \;\; \texttt{correct}_i = 1 \right\},
    \end{equation}
    
    where $\texttt{correct}_i$ indicates whether the agent's final answer in trace $i$ is correct.

To measure recovery behavior, we introduce the Evidence Recovery Function (ERF), which quantifies the cumulative proportion of traces that have successfully recovered by each turn:

\begin{definition}[Evidence Recovery Function (ERF)]\label{def:erf}
Let $N$ denote the total number of traces. The \emph{Evidence Recovery Function} at turn $t$ is defined as
\begin{equation}\label{eq:erf_evidence}
\mathrm{ERF}(t) := \frac{1}{N} \sum_{i=1}^{N} \mathbb{I}\left(T_{\text{recover},i} \leq t\right),
\end{equation}
where $\mathbb{I}(\cdot)$ is the indicator function. $\mathrm{ERF}(t)$ measures the proportion of traces that have recovered by turn $t$.
\end{definition}

\subsubsection{Calibration}\label{sec:o3_calibration}

We evaluate \textbf{calibration} as the agent's ability to decide \emph{when to answer} based on the quality of retrieved evidence. A well-calibrated agent should answer only when it has acquired evidence that is both clear (unambiguous and directly relevant) and sufficient (contains enough information to support a reliable answer).

Let $\texttt{answer}_{i,t} \in \{0,1\}$ indicate whether the agent provides an answer at turn $t$ of trace $i$. We assess calibration behavior by examining the answer rate conditioned on the evidence state:
\begin{equation}
\mathbb{P}(\texttt{answer}_{i,t} = 1 \mid E_{i,t} = k).
\label{eq:step-calib-answer}
\end{equation}

High values at $k=0$ indicate \emph{epistemic overconfidence}, where the agent answers prematurely with poor or partial evidence. Conversely, low values at $k=2$ suggest \emph{epistemic overcautiousness}, where the agent refrains from answering when the evidence is good.

To quantify calibration performance, we introduce \textbf{Calibration Error (CE)} that measures how much an agent's answering behavior deviates from the ideal policy. The ideal policy is one that answers if and only if the evidence is good ($E_{i,t} = 2$), which maximizes expected accuracy while minimizing wasted effort. This metric captures both epistemic failures: overconfidence (answering with insufficient evidence) and overcautiousness (not answering despite having good evidence).

\begin{definition}[Calibration Error (CE)]
Let $\mathcal{I} := \{1, \dots, N\}$ be the index set of traces. Let $\pi^*(k) := \mathbb{I}[k=2]$ represent the ideal policy that answers if and only if evidence is sufficient. The CE for a model is defined as:
\begin{equation}\label{eq:step-calib-ce}
\mathrm{CE} := \mathbb{E}_{i \in \mathcal{I}} [\mathrm{CE}_i]
\end{equation}
where for each trace $i$, \(
\mathrm{CE}_i \!:=\! \sum_{k=0}^{2}\! \mathbb{P}(E_{i,t}\!=\!k) \left| \mathbb{P}(\texttt{answer}_{i,t}=1 \mid E_{i,t}=k) \!-\! \pi^*(k) \right|
\).
\end{definition}
For a perfectly calibrated agent following the ideal policy $\pi^*(k)$, it achieves $\mathrm{CE} = 0$.

\section{Experiments}

\subsection{Experimental Setup}

\paragraph{Models \& Datasets}. We evaluate Qwen-2.5-7B-Instruct (``Base'' for training, \cite{qwen2024qwen2}), its few-shot prompted version (``Few-shot''), and state-of-the-art ``RL-trained'' agents, including: \textsc{Search-R1} \citep{searchr1}, \textsc{ReSearch} \citep{research}, \textsc{ASearcher} \citep{asearcher} and \textsc{DeepResearcher} \citep{deepresearcher}. 
We evaluate the agents on a diverse set of seven question-answering benchmarks: NQ~\citep{nq}, TriviaQA~\citep{triviaqa}, and PopQA~\citep{popqa} [\textcolor{blue}{single-hop}]; HotpotQA~\citep{hotpotqa}, 2Wiki~\citep{2wikimultihopqa}, MusiQue~\citep{musique}, and Bamboogle~\citep{bamboogle} [\textcolor{blue}{multi-hop}]. 

Each model runs and evaluates on the \textbf{sanitized} test datasets, where we remove ambiguous questions and data contamination cases to ensure evaluation quality (see Appendix~\ref{app:data_sanitization} for details).
Our evaluation comprises \textbf{28,493} traces and \textbf{283,950} steps across all models and datasets.
For this large-scale annotation, we employ GPT-4.1-mini, which demonstrates substantial alignment with human judgments under our validated schema.

\paragraph{Answer-level performance.}
Aggregate F1 ranking is \textsc{ASearcher} $>$ Search-R1 $>$ \textsc{ReSearch} $>$ Few-shot $\approx$ \textsc{DeepResearcher} $>$ Base (Appendix~\ref{app:accuracy}).
Our primary analysis moves beyond outcomes to assess \emph{process-level epistemic competencies}: evidence grounding (\Cref{sec:o1_reasoning}), recovery dynamics (\Cref{sec:exp_recovery}), and calibration (\Cref{sec:exp_calibration}).
We identify agent-specific competencies that drive performance gains and expose the overestimation of RL training (\Cref{sec:exp_agent_usage}).

\subsection{Evaluating Justified Reasoning via Evidence Grounding}\label{sec:o1_reasoning}

A fundamental criterion of agent competence is not merely producing the correct answer, but doing so through a reasoning process \emph{explicitly grounded in retrieved evidence}.
To measure this, we utilize the Reasoning Quality Index (RQI, defined in~\Cref{sec:o1_groundness}), which quantifies the proportion of an agent's reasoning steps that are supported by retrieved evidence.

\begin{figure}[t] 
    \centering 
    \includegraphics[width=0.95\linewidth]{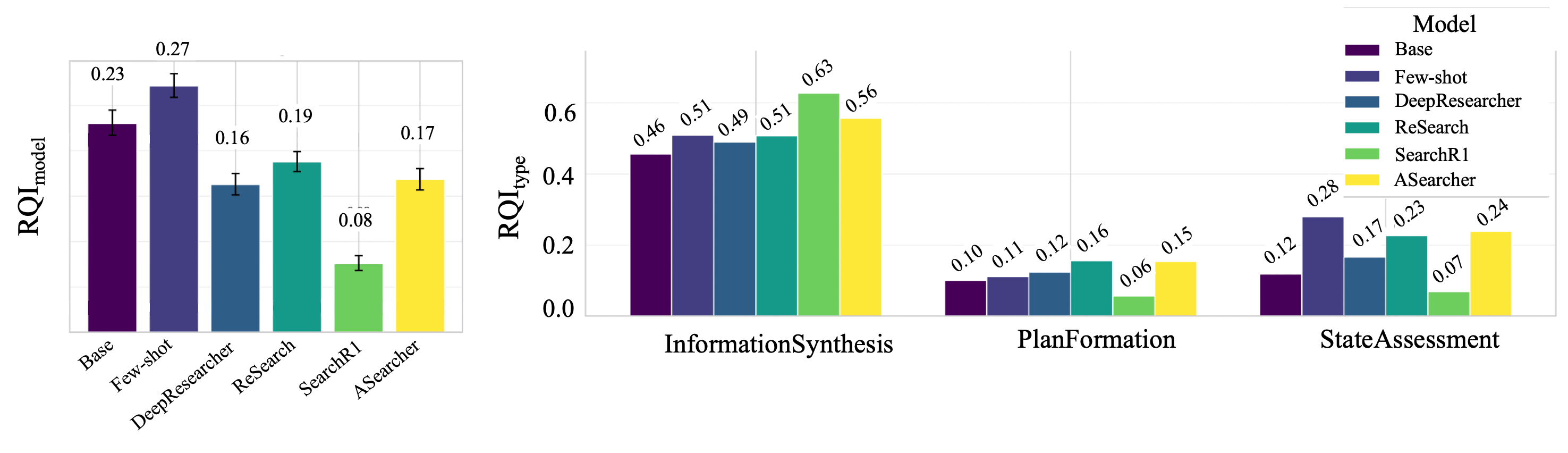} 
    \caption{\textbf{RQI Analysis Summary.} \emph{Left:} RQI by model level, showing the overall reasoning quality across different agent types. \emph{Right:} RQI by reasoning type, revealing that models struggle most with plan formation and state assessment compared to information synthesis.} 
    \label{fig:rqi_summary}
    \end{figure}

\paragraph{RL Training Fails to Develop Evidence-Grounded Reasoning.}
Figure~\ref{fig:rqi_summary} (\emph{Left}) presents the average RQI scores across models. Few-shot prompting achieves the highest reasoning quality (RQI = 0.27), outperforming all RL-trained agents. This reveals a \emph{disconnect between answer-level success and reasoning groundedness}: RL training may optimize for correct final answers, but it fails to develop the epistemic reasoning skills needed to justify those answers through evidence-grounded reasoning.

\paragraph{Plan Formation and State Assessment Are Core Reasoning Failures.}
To understand \emph{where} reasoning breaks down, we analyze performance by reasoning type (Figure~\ref{fig:rqi_summary}, \emph{Right}). Specifically,
\begin{itemize}
    \item \textbf{Information Synthesis} emerges as a relative strength across models (e.g., \textsc{ASearcher}: 0.56), demonstrating agents' proficiency in summarizing and restating retrieved information.
    \item \textbf{Plan Formation} constitutes the most significant weakness for all agents (consistently scoring below 0.2), highlighting fundamental difficulties in breaking down complex queries and formulating coherent search strategies.
    \item \textbf{State Assessment} shows notable improvement in few-shot models (0.28), suggesting enhanced metacognitive capabilities compared to their RL-trained counterparts.
\end{itemize}

For detailed analysis of evidence-conditioned reasoning quality across different evidence states and reasoning types, see Appendix~\ref{app:evidence_grounded_reasoning}.

\subsection{Recovery Analysis}\label{sec:exp_recovery}
\begin{figure}[t]
    \centering
    \includegraphics[width=0.95\linewidth]{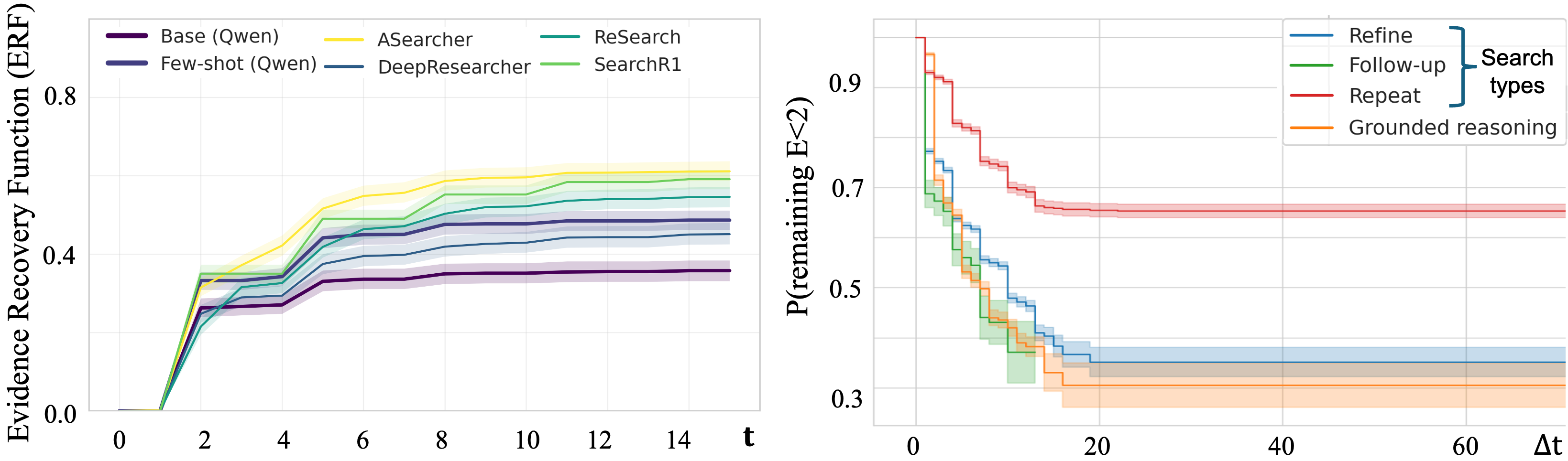}
    \caption{\small \textbf{Recovery Analysis.} \emph{Left:} ERFs by model showing recovery from low to sufficient evidence (\texttt{E=2}) as turn $t$ increases. \emph{Right:} Recovery efficiency by action type. Steeper curves indicate faster escape from low evidence states. \textsc{Refine} and \textsc{Follow-up} enable fastest recovery, while \textsc{Repeat} shows minimal improvement.}
    \label{fig:recovery_analysis}
\end{figure}

This section evaluates whether models can effectively recover from low-quality evidence through adaptive search strategies.
\paragraph{Recovery Competence with ERF.}
We first assess overall recovery competence using the Evidence Recovery Function (ERF, \Cref{eq:erf_evidence}), which measures the cumulative probability of reaching sufficient evidence ($E=2$) over time. As shown in \Cref{fig:recovery_analysis} (\emph{Left}), \textsc{ASearcher}, which has the highest F1 score on answer correctness, shows superior recovery performance compared to other agents. 
In contrast, \textsc{DeepResearcher}, which has the lowest F1 among all RL-trained agents, shows the poorest recovery performance.
This demonstrates that \emph{effective algorithm design should prioritize developing adaptive evidence-seeking strategies so that agents can recover from insufficient evidence and improve final performance}.
\paragraph{Refine and Follow-up Search Strategies Drive Effective Recovery.}
To identify the \emph{most effective search strategies} for recovery, we analyze how different action types affect recovery rates over time. 
We categorize all search and reasoning steps by their types. 
For each step $t$ of a specific type in trace, we measure the proportion of turns remaining low evidence states ($E<2$) at subsequent turns ($t + \Delta t$). 
Given the variable-length traces and resulting right-censored data (traces ending before recovery occurs—when observation periods end before the outcome), we employ Kaplan-Meier survival analysis \citep{kaplan1958nonparametric}, which provides robust estimation of recovery probabilities despite incomplete observations.

As shown in \Cref{fig:recovery_analysis} (\emph{Right}), survival curves reveal that \textsc{Refine} and \textsc{Follow-up} strategies enable the fastest recovery from low-quality evidence, while \textsc{Repeat} provides minimal benefit. Additionally, \textsc{Grounded reasoning} also effectively improves evidence utilization in responses.

% Overall, \emph{strategic action selection is crucial for recovery}—agents must learn to avoid repetitive, unproductive queries and instead employ exploratory search strategies that integrate new information with existing knowledge.

\subsection{Evidence-Aligned Calibration}\label{sec:exp_calibration}
\begin{wrapfigure}{r}{0.5\linewidth}
    \centering
    \includegraphics[width=\linewidth]{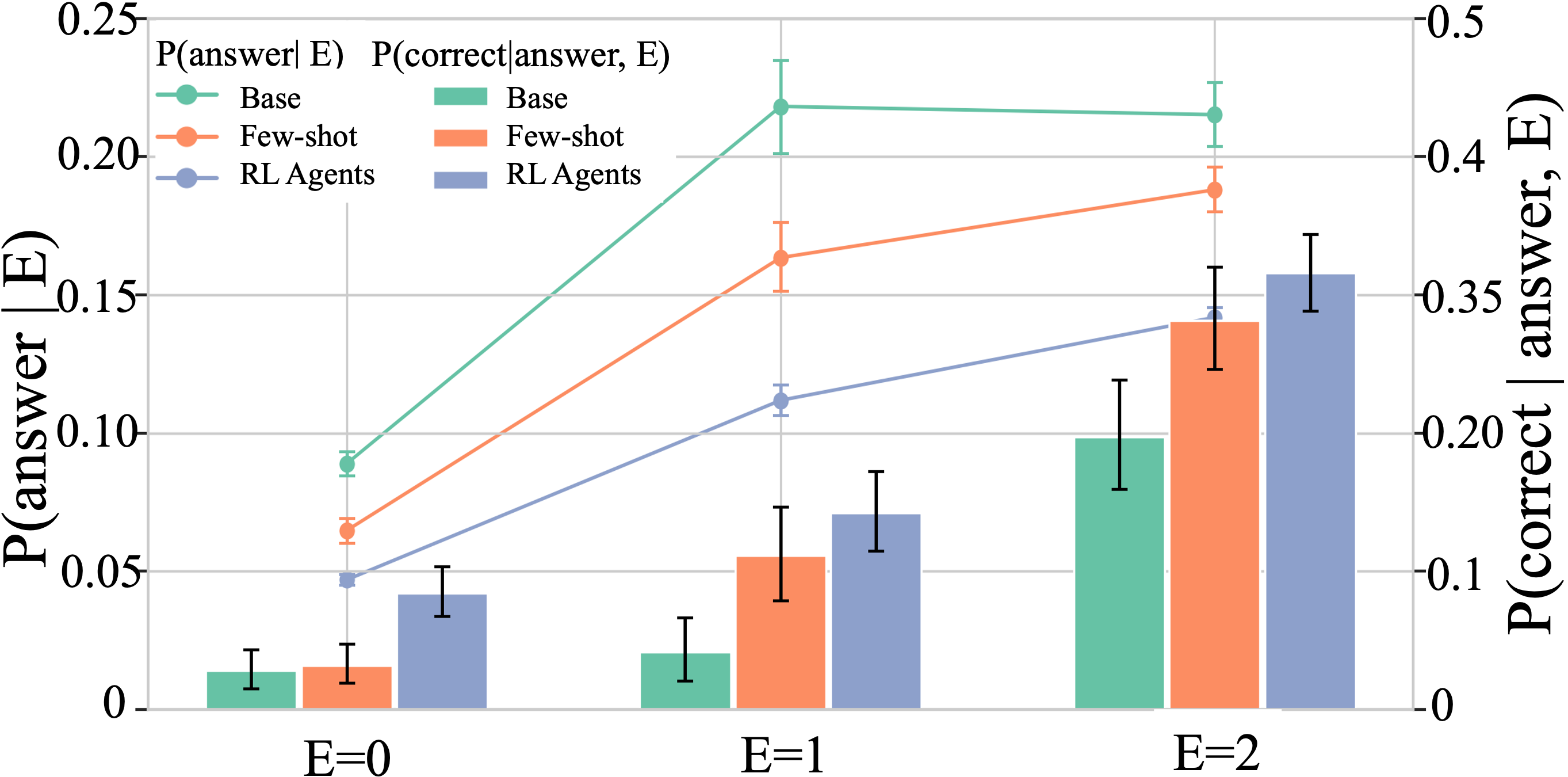}
    \caption{
    \textbf{Evidence State Drives Answer Probability and Accuracy.}  
    \emph{Lines:} Answer probability increases with evidence state.
    \emph{Bars:} Answer accuracy improves with evidence state.
    RL-trained models show lower answering rate but higher accuracy with good evidence.\looseness=-1
    }
    \label{fig:r3_evidence_summary}
  \end{wrapfigure}
  \par\mbox{} 
This section evaluates whether models calibrate their answering behavior to the ideal policy, where they answer when and only when it has good evidence, avoiding both overconfidence (answering with poor evidence) and overcautiousness (failing to answer despite good evidence).
\paragraph{Evidence Quality Drives Answer Accuracy.}
We first validate that evidence quality correlates with answer accuracy. As shown in Figure~\ref{fig:r3_evidence_summary}, RL-trained models achieve $31.6\%$ accuracy when answering with good evidence ($P(\text{correct}|\text{answer},E=2)$), compared to only $8.4\%$ accuracy when answering without supporting evidence. This significant difference demonstrates that \emph{evidence quality is positively associated with answer correctness}.

Interestingly, RL-trained models exhibit lower answering rates ($P(\text{answer}|E)$, formally defined in Equation~\ref{eq:step-calib-answer}) across all evidence states compared to base models.
This suggests that RL training encourages models to be \emph{more selective about when to provide final answers}, potentially reducing instances of overconfident responses.

To further understand the calibration behavior, we measure calibration quality using \textbf{calibration error} (CE, \Cref{eq:step-calib-ce}) and analyze two specific failure modes to identify where models fail:\looseness=-1
\begin{itemize}[leftmargin=2em,itemsep=0pt,parsep=0pt,topsep=0pt]
    \item[(1)] \textbf{Overconfident answering}: providing a final answer when the trace never reached good evidence state ($E_{i,t} < 2$ for all $t$), indicating overconfidence;
    \item[(2)] \textbf{Overcautious abstention}: failing to provide a final answer despite having reached good evidence state ($E_{i,t} = 2$), indicating underconfidence.
\end{itemize}
\paragraph{RL Training Improves Calibration.}
As summarized in \Cref{tab:calibration_error_rates}, RL-trained models show substantial improvements in calibration behavior. They reduce overconfident answering from $63.1\%$ to $35.3\%$ and achieve the lowest overall calibration error ($0.309$). This demonstrates that RL training successfully teaches models to align their answering decisions with evidence quality, moving toward the ideal policy of answering only when evidence is sufficient. This finding contrasts with the earlier result that RL training degrades reasoning groundedness (Section~\ref{sec:o1_reasoning}), highlighting the \emph{competency-specific nature} of RL training effects. For detailed analysis of individual RL-trained agents and evidence-conditioned answer timing patterns, see Appendix~\ref{app:calibration_model_breakdown}.

\begin{table}[t]
\centering
\caption{
\textbf{Calibration Error Analysis.}  
traces categorized as: calibration error, overconfident, or overcautious. Lower values indicate better calibration. RL-trained agents show the lowest overconfident answer rate and lowest CE. \textbf{Bold} indicates best performance.\looseness=-1
}
\label{tab:calibration_error_rates}
\small
\begin{tabular}{lccc}
\toprule
\textbf{Model} & \textbf{(1) Overconfident} $\downarrow$ & \textbf{(2) Overcautious} $\downarrow$ & \textbf{(3) Calibration Error} $\downarrow$ \\
\midrule
Base      & 0.631 & {0.030} & {0.329} \\
Few-shot  & 0.511 & \textbf{0.024} & 0.317 \\
RL-trained   & \textbf{0.353} & 0.085 & \textbf{0.309} \\
\bottomrule
\end{tabular}
\end{table}
\subsection{Exploiting Epistemic Competencies for Performance Gains}\label{sec:exp_agent_usage}

Our evaluation reveals distinct agent specializations: \textsc{ASearcher} excels in \emph{evidence acquisition} and \emph{recovery mechanisms} (highest overall F1 score), while \textsc{Search-R1} demonstrates superior \textbf{information synthesis} (RQI=0.63 for information synthesis) with \textbf{minimal overconfident answering} (see \Cref{sec:exp_calibration} and Appendix~\ref{app:calibration_model_breakdown}). These findings motivated us to explore \emph{agent synthesis}—using one agent's evidence collection as input for another's answer generation.

We provided agents with reasoning traces and evidence from others, then measured F1 score improvements. \textsc{Search-R1} emerges as the \textbf{most effective synthesizer} (+1.27 F1 on average), significantly outperforming other agents (see details in~Appendix~\ref{app:agent_synthesis}). 
Surprisingly, Base achieved the highest F1 gains (+2.42 on average) when paired with other models for answer generation. This reveals that accuracy-only evaluation may \textbf{underestimate} Base’s reasoning abilities while overstating the gains from RL training.

Our method reveals distinct agent profiles by systematically benchmarking their epistemic competencies, for example, \textsc{Search-R1}'s synthesis strength and conservative answering. These insights provide a reliable foundation for designing effective systems that capitalize on complementary agent strengths. Overall, beyond outcome-based metrics, our approach delivers procedural evaluation that enables more interpretable assessments of agent competence.
\

\section{Conclusion}
\vspace{-0.5em}
\algo evaluates epistemic competence in LLM search agents through expert-annotated traces, revealing gaps in current evaluation approaches. Our evidence state framework and metrics (RQI, ERF, CE) show that RL training improves answer accuracy but not evidence-grounded reasoning. Our framework uncovers agent-specific strengths masked by accuracy-only evaluation: Search-R1 excels at evidence synthesis, while Base models demonstrate stronger reasoning capabilities than accuracy metrics suggest. This work establishes epistemic competence as essential for reliable AI systems and provides a framework for developing reliable information-seeking agents. Future work should explore modular architectures combining complementary strengths and training approaches that improve higher-order reasoning alongside answer calibration.

\newpage
\bibliographystyle{plainnat}
\bibliography{ref}

\begin{thebibliography}{40}
\providecommand{\natexlab}[1]{#1}
\providecommand{\url}[1]{\texttt{#1}}
\expandafter\ifx\csname urlstyle\endcsname\relax
  \providecommand{\doi}[1]{doi: #1}\else
  \providecommand{\doi}{doi: \begingroup \urlstyle{rm}\Url}\fi

\bibitem[Chen et~al.(2025)Chen, Li, Sun, Zhou, Zhu, Wang, Pan, Zhang, Chen,
  Yang, et~al.]{research}
Mingyang Chen, Tianpeng Li, Haoze Sun, Yijie Zhou, Chenzheng Zhu, Haofen Wang,
  Jeff~Z Pan, Wen Zhang, Huajun Chen, Fan Yang, et~al.
\newblock Learning to reason with search for llms via reinforcement learning.
\newblock \emph{arXiv preprint arXiv:2503.19470}, 2025.

\bibitem[Coeckelbergh(2023)]{coeckelbergh2023democracy}
Mark Coeckelbergh.
\newblock Democracy, epistemic agency, and ai: political epistemology in times
  of artificial intelligence.
\newblock \emph{AI and Ethics}, 3\penalty0 (4):\penalty0 1341--1350, 2023.

\bibitem[Cohen(1960)]{cohen1960coefficient}
Jacob Cohen.
\newblock A coefficient of agreement for nominal scales.
\newblock \emph{Educational and psychological measurement}, 20\penalty0
  (1):\penalty0 37--46, 1960.

\bibitem[Cronbach and Meehl(1955)]{cronbach1955construct}
Lee~J Cronbach and Paul~E Meehl.
\newblock Construct validity in psychological tests.
\newblock \emph{Psychological bulletin}, 52\penalty0 (4):\penalty0 281, 1955.

\bibitem[Gao et~al.(2025)Gao, Fu, Xie, Xu, He, Mei, Zhu, and Wu]{asearcher}
Jiaxuan Gao, Wei Fu, Minyang Xie, Shusheng Xu, Chuyi He, Zhiyu Mei, Banghua
  Zhu, and Yi~Wu.
\newblock Beyond ten turns: Unlocking long-horizon agentic search with
  large-scale asynchronous rl.
\newblock \emph{arXiv preprint arXiv:2508.07976}, 2025.

\bibitem[Greene et~al.(2016)Greene, Sandoval, and
  Br{\aa}ten]{greene2016introduction}
Jeffrey~A Greene, William~A Sandoval, and Ivar Br{\aa}ten.
\newblock An introduction to epistemic cognition.
\newblock In \emph{Handbook of epistemic cognition}, pages 1--16. Routledge,
  2016.

\bibitem[Guo et~al.(2025)Guo, Yang, Zhang, Song, Zhang, Xu, Zhu, Ma, Wang, Bi,
  et~al.]{guo2025deepseek}
Daya Guo, Dejian Yang, Haowei Zhang, Junxiao Song, Ruoyu Zhang, Runxin Xu,
  Qihao Zhu, Shirong Ma, Peiyi Wang, Xiao Bi, et~al.
\newblock Deepseek-r1: Incentivizing reasoning capability in llms via
  reinforcement learning.
\newblock \emph{arXiv preprint arXiv:2501.12948}, 2025.

\bibitem[Ho et~al.(2020)Ho, Nguyen, Sugawara, and Aizawa]{2wikimultihopqa}
Xanh Ho, Anh-Khoa~Duong Nguyen, Saku Sugawara, and Akiko Aizawa.
\newblock Constructing a multi-hop qa dataset for comprehensive evaluation of
  reasoning steps.
\newblock \emph{arXiv preprint arXiv:2011.01060}, 2020.

\bibitem[Jaech et~al.(2024)Jaech, Kalai, Lerer, Richardson, El-Kishky, Low,
  Helyar, Madry, Beutel, Carney, et~al.]{jaech2024openai}
Aaron Jaech, Adam Kalai, Adam Lerer, Adam Richardson, Ahmed El-Kishky, Aiden
  Low, Alec Helyar, Aleksander Madry, Alex Beutel, Alex Carney, et~al.
\newblock Openai o1 system card.
\newblock \emph{arXiv preprint arXiv:2412.16720}, 2024.

\bibitem[Jin et~al.(2025)Jin, Zeng, Yue, Yoon, Arik, Wang, Zamani, and
  Han]{searchr1}
Bowen Jin, Hansi Zeng, Zhenrui Yue, Jinsung Yoon, Sercan Arik, Dong Wang, Hamed
  Zamani, and Jiawei Han.
\newblock Search-r1: Training llms to reason and leverage search engines with
  reinforcement learning.
\newblock \emph{arXiv preprint arXiv:2503.09516}, 2025.

\bibitem[Joshi et~al.(2017)Joshi, Choi, Weld, and Zettlemoyer]{triviaqa}
Mandar Joshi, Eunsol Choi, Daniel~S Weld, and Luke Zettlemoyer.
\newblock Triviaqa: A large scale distantly supervised challenge dataset for
  reading comprehension.
\newblock \emph{arXiv preprint arXiv:1705.03551}, 2017.

\bibitem[Kaplan and Meier(1958)]{kaplan1958nonparametric}
Edward~L Kaplan and Paul Meier.
\newblock Nonparametric estimation from incomplete observations.
\newblock \emph{Journal of the American statistical association}, 53\penalty0
  (282):\penalty0 457--481, 1958.

\bibitem[Krippendorff(2018)]{krippendorff2018content}
Klaus Krippendorff.
\newblock \emph{Content analysis: An introduction to its methodology}.
\newblock Sage publications, 2018.

\bibitem[Kumar et~al.(2025)Kumar, Yu, Hallman, Pokrass, Goucher, Ganesh, Cheng,
  McKinzie, Zhang, Koch, Wei, Medina, Wong, Kavanaugh, Bekerman, Hu, Ren,
  Singal, Kiros, Ai, et~al.]{openai2025gpt4_1}
Ananya Kumar, Jiahui Yu, John Hallman, Michelle Pokrass, Adam Goucher, Adi
  Ganesh, Bowen Cheng, Brandon McKinzie, Brian Zhang, Chris Koch, Colin Wei,
  David Medina, Edmund Wong, Erin Kavanaugh, Florent Bekerman, Haitang Hu,
  Hongyu Ren, Ishaan Singal, Jamie Kiros, Jason Ai, et~al.
\newblock Introducing gpt-4.1 in the api.
\newblock \url{https://openai.com/index/gpt-4-1/}, 2025.

\bibitem[Kwiatkowski et~al.(2019)Kwiatkowski, Palomaki, Redfield, Collins,
  Parikh, Alberti, Epstein, Polosukhin, Devlin, Lee, et~al.]{nq}
Tom Kwiatkowski, Jennimaria Palomaki, Olivia Redfield, Michael Collins, Ankur
  Parikh, Chris Alberti, Danielle Epstein, Illia Polosukhin, Jacob Devlin,
  Kenton Lee, et~al.
\newblock Natural questions: a benchmark for question answering research.
\newblock \emph{Transactions of the Association for Computational Linguistics},
  7:\penalty0 453--466, 2019.

\bibitem[Li et~al.(2025)Li, Zhang, Yin, Zhang, Ou, Wu, Yin, Li, Tao, Wang,
  et~al.]{websailor}
Kuan Li, Zhongwang Zhang, Huifeng Yin, Liwen Zhang, Litu Ou, Jialong Wu,
  Wenbiao Yin, Baixuan Li, Zhengwei Tao, Xinyu Wang, et~al.
\newblock Websailor: Navigating super-human reasoning for web agent.
\newblock \emph{arXiv preprint arXiv:2507.02592}, 2025.

\bibitem[Li et~al.(2024)Li, Wang, Tran, Xia, and Du]{meqa}
Ruosen Li, Zimu Wang, Son Tran, Lei Xia, and Xinya Du.
\newblock Meqa: A benchmark for multi-hop event-centric question answering with
  explanations.
\newblock \emph{Advances in Neural Information Processing Systems},
  37:\penalty0 126835--126862, 2024.

\bibitem[Mallen et~al.(2022)Mallen, Asai, Zhong, Das, Khashabi, and
  Hajishirzi]{popqa}
Alex Mallen, Akari Asai, Victor Zhong, Rajarshi Das, Daniel Khashabi, and
  Hannaneh Hajishirzi.
\newblock When not to trust language models: Investigating effectiveness of
  parametric and non-parametric memories.
\newblock \emph{arXiv preprint arXiv:2212.10511}, 2022.

\bibitem[Maohao~Shen et~al.(2025)Maohao~Shen, Zhenting~Qi, Hong, Wei~Lu,
  Subhro~Das, and Gan]{satori25}
Guangtao~Zeng Maohao~Shen, Zhang-Wei Zhenting~Qi, Zhenfang~Chen Hong,
  Gregory~Wornell Wei~Lu, David~Cox Subhro~Das, and Chuang Gan.
\newblock Satori: Reinforcement learning with chain-of-action-thought enhances
  llm reasoning via autoregressive search, 2025.
\newblock URL \url{https://arxiv.org/abs/2502.02508}.

\bibitem[Mukherjee et~al.(2025)Mukherjee, Chinta, Kim, Sharma, and
  Tur]{mukherjee2025premiseaugmented}
Sagnik Mukherjee, Abhinav Chinta, Takyoung Kim, Tarun~Anoop Sharma, and
  Dilek~Hakkani Tur.
\newblock Premise-augmented reasoning chains improve error identification in
  math reasoning with {LLM}s.
\newblock In \emph{Forty-second International Conference on Machine Learning},
  2025.
\newblock URL \url{https://openreview.net/forum?id=4tYckHNVXV}.

\bibitem[Nguyen et~al.(2024)Nguyen, Luo, Shiri, Phung, Li, Vu, and
  Haffari]{nguyen2024direct}
Minh-Vuong Nguyen, Linhao Luo, Fatemeh Shiri, Dinh Phung, Yuan-Fang Li,
  Thuy-Trang Vu, and Gholamreza Haffari.
\newblock Direct evaluation of chain-of-thought in multi-hop reasoning with
  knowledge graphs.
\newblock \emph{arXiv preprint arXiv:2402.11199}, 2024.

\bibitem[OpenAI(2025)]{openai2025gpt5}
OpenAI.
\newblock Introducing gpt-5.
\newblock \url{https://openai.com/index/introducing-gpt-5/}, 2025.

\bibitem[Ott et~al.(2023)Ott, Hebenstreit, Liévin, Hother, Moradi, Mayrhauser,
  Praas, Winther, and Samwald]{Ott2023}
Simon Ott, Konstantin Hebenstreit, Valentin Liévin, Christoffer~Egeberg
  Hother, Milad Moradi, Maximilian Mayrhauser, Robert Praas, Ole Winther, and
  Matthias Samwald.
\newblock Thoughtsource: A central hub for large language model reasoning data.
\newblock \emph{Scientific Data}, 10\penalty0 (1), August 2023.
\newblock ISSN 2052-4463.
\newblock \doi{10.1038/s41597-023-02433-3}.
\newblock URL \url{http://dx.doi.org/10.1038/s41597-023-02433-3}.

\bibitem[Paul et~al.(2024)Paul, West, Bosselut, and Faltings]{paul2024making}
Debjit Paul, Robert West, Antoine Bosselut, and Boi Faltings.
\newblock Making reasoning matter: Measuring and improving faithfulness of
  chain-of-thought reasoning.
\newblock \emph{arXiv preprint arXiv:2402.13950}, 2024.

\bibitem[Press et~al.(2022)Press, Zhang, Min, Schmidt, Smith, and
  Lewis]{bamboogle}
Ofir Press, Muru Zhang, Sewon Min, Ludwig Schmidt, Noah~A Smith, and Mike
  Lewis.
\newblock Measuring and narrowing the compositionality gap in language models.
\newblock \emph{arXiv preprint arXiv:2210.03350}, 2022.

\bibitem[Qwen et~al.(2024)Qwen, Yang, Zhang, Hui, Zheng, Yu, Li, Liu, Huang,
  Wei, et~al.]{qwen2024qwen2}
A~Yang Qwen, Baosong Yang, B~Zhang, B~Hui, B~Zheng, B~Yu, Chengpeng Li, D~Liu,
  F~Huang, H~Wei, et~al.
\newblock Qwen2. 5 technical report.
\newblock \emph{arXiv preprint}, 2024.

\bibitem[Shi et~al.(2025)Shi, Li, Wu, Liu, Fang, Cai, Zhang, and
  Wang]{autorefine}
Yaorui Shi, Sihang Li, Chang Wu, Zhiyuan Liu, Junfeng Fang, Hengxing Cai,
  An~Zhang, and Xiang Wang.
\newblock Search and refine during think: Autonomous retrieval-augmented
  reasoning of llms.
\newblock \emph{arXiv preprint arXiv:2505.11277}, 2025.

\bibitem[Song et~al.(2025)Song, Jiang, Min, Chen, Chen, Zhao, Fang, and
  Wen]{r1searcher}
Huatong Song, Jinhao Jiang, Yingqian Min, Jie Chen, Zhipeng Chen, Wayne~Xin
  Zhao, Lei Fang, and Ji-Rong Wen.
\newblock R1-searcher: Incentivizing the search capability in llms via
  reinforcement learning.
\newblock \emph{arXiv preprint arXiv:2503.05592}, 2025.

\bibitem[Trivedi et~al.(2022)Trivedi, Balasubramanian, Khot, and
  Sabharwal]{musique}
Harsh Trivedi, Niranjan Balasubramanian, Tushar Khot, and Ashish Sabharwal.
\newblock Musique: Multihop questions via single-hop question composition.
\newblock \emph{Transactions of the Association for Computational Linguistics},
  10:\penalty0 539--554, 2022.

\bibitem[Wei et~al.(2022)Wei, Wang, Schuurmans, Bosma, Xia, Chi, Le, Zhou,
  et~al.]{wei2022chain}
Jason Wei, Xuezhi Wang, Dale Schuurmans, Maarten Bosma, Fei Xia, Ed~Chi, Quoc~V
  Le, Denny Zhou, et~al.
\newblock Chain-of-thought prompting elicits reasoning in large language
  models.
\newblock \emph{Advances in neural information processing systems},
  35:\penalty0 24824--24837, 2022.

\bibitem[Wu et~al.(2025)Wu, Cai, Mei, Wen, Hu, Yang, Fu, and
  Shi]{wu2025kgtraces}
Rong Wu, Pinlong Cai, Jianbiao Mei, Licheng Wen, Tao Hu, Xuemeng Yang, Daocheng
  Fu, and Botian Shi.
\newblock Kg-traces: Enhancing large language models with knowledge
  graph-constrained trajectory reasoning and attribution supervision, 2025.
\newblock URL \url{https://arxiv.org/abs/2506.00783}.

\bibitem[Xi et~al.(2025{\natexlab{a}})Xi, Lin, Xiao, Zhou, Shan, Gao, Zhu, Liu,
  Yu, and Zhang]{xi2025survey}
Yunjia Xi, Jianghao Lin, Yongzhao Xiao, Zheli Zhou, Rong Shan, Te~Gao, Jiachen
  Zhu, Weiwen Liu, Yong Yu, and Weinan Zhang.
\newblock A survey of llm-based deep search agents: Paradigm, optimization,
  evaluation, and challenges.
\newblock \emph{arXiv preprint arXiv:2508.05668}, 2025{\natexlab{a}}.

\bibitem[Xi et~al.(2025{\natexlab{b}})Xi, Lin, Zhu, Xiao, Ou, Liu, Wan, Chen,
  Liu, Wang, et~al.]{infodeepseek}
Yunjia Xi, Jianghao Lin, Menghui Zhu, Yongzhao Xiao, Zhuoying Ou, Jiaqi Liu,
  Tong Wan, Bo~Chen, Weiwen Liu, Yasheng Wang, et~al.
\newblock Infodeepseek: Benchmarking agentic information seeking for
  retrieval-augmented generation.
\newblock \emph{arXiv preprint arXiv:2505.15872}, 2025{\natexlab{b}}.

\bibitem[Xiong et~al.(2025)Xiong, Cai, Li, and Wang]{xiong2025mapping}
Zhen Xiong, Yujun Cai, Zhecheng Li, and Yiwei Wang.
\newblock Mapping the minds of llms: A graph-based analysis of reasoning llm.
\newblock \emph{arXiv preprint arXiv:2505.13890}, 2025.

\bibitem[Yang et~al.(2018)Yang, Qi, Zhang, Bengio, Cohen, Salakhutdinov, and
  Manning]{hotpotqa}
Zhilin Yang, Peng Qi, Saizheng Zhang, Yoshua Bengio, William~W Cohen, Ruslan
  Salakhutdinov, and Christopher~D Manning.
\newblock Hotpotqa: A dataset for diverse, explainable multi-hop question
  answering.
\newblock \emph{arXiv preprint arXiv:1809.09600}, 2018.

\bibitem[Yao et~al.(2023)Yao, Zhao, Yu, Du, Shafran, Narasimhan, and
  Cao]{yao2023react}
Shunyu Yao, Jeffrey Zhao, Dian Yu, Nan Du, Izhak Shafran, Karthik Narasimhan,
  and Yuan Cao.
\newblock React: Synergizing reasoning and acting in language models.
\newblock In \emph{International Conference on Learning Representations
  (ICLR)}, 2023.

\bibitem[Zhang et~al.(2025{\natexlab{a}})Zhang, Geng, Yu, Yin, Zhang, Tan,
  Zhou, Li, Xue, Li, et~al.]{zhang2025landscape}
Guibin Zhang, Hejia Geng, Xiaohang Yu, Zhenfei Yin, Zaibin Zhang, Zelin Tan,
  Heng Zhou, Zhongzhi Li, Xiangyuan Xue, Yijiang Li, et~al.
\newblock The landscape of agentic reinforcement learning for llms: A survey.
\newblock \emph{arXiv preprint arXiv:2509.02547}, 2025{\natexlab{a}}.

\bibitem[Zhang et~al.(2023)Zhang, Li, Cui, Cai, Liu, Fu, Huang, Zhao, Zhang,
  Chen, et~al.]{zhang2023siren}
Yue Zhang, Yafu Li, Leyang Cui, Deng Cai, Lemao Liu, Tingchen Fu, Xinting
  Huang, Enbo Zhao, Yu~Zhang, Yulong Chen, et~al.
\newblock Siren's song in the ai ocean: a survey on hallucination in large
  language models.
\newblock \emph{arXiv preprint arXiv:2309.01219}, 2023.

\bibitem[Zhang et~al.(2025{\natexlab{b}})Zhang, Yang, Shu, Wen, and
  Sang]{autocoa}
Yuxiang Zhang, Yuqi Yang, Jiangming Shu, Xinyan Wen, and Jitao Sang.
\newblock Agent models: Internalizing chain-of-action generation into reasoning
  models.
\newblock \emph{arXiv preprint arXiv:2503.06580}, 2025{\natexlab{b}}.

\bibitem[Zheng et~al.(2025)Zheng, Fu, Hu, Cai, Ye, Lu, and Liu]{deepresearcher}
Yuxiang Zheng, Dayuan Fu, Xiangkun Hu, Xiaojie Cai, Lyumanshan Ye, Pengrui Lu,
  and Pengfei Liu.
\newblock Deepresearcher: Scaling deep research via reinforcement learning in
  real-world environments.
\newblock \emph{arXiv preprint arXiv:2504.03160}, 2025.

\end{thebibliography}
{\small\tableofcontents}
\newpage

\section{Data Sanitization}
\label{app:data_sanitization}
To ensure the quality of our evaluation, we sanitize the test sets of our seven benchmark datasets using the following two criteria:

(1) \textbf{Ambiguous or Unanswerable Questions.} We discard questions where benchmark agents receive full credit for speculative answers, while a stronger reference model (GPT-4.1-mini) abstains with a justified explanation. For example, in response to the question \textit{“Who developed the CPU?”}, a benchmark agent might confidently output \textit{“John von Neumann”}, achieving EM=1. In contrast, GPT-4.1-mini responds:
\begin{scriptsize}
  ``\texttt{Answer: I don't know};
\texttt{Reason: The information mentions figures like John von Neumann and J. Presper Eckert, but does not identify a single developer.}''  
\end{scriptsize}
These questions are excluded to avoid rewarding superficial matching over careful reasoning.

(2) \textbf{Data Contamination.} We discard questions where agents succeed (Pass@3) without issuing any search queries, as this indicates the question is likely part of the model's pre-training data. These are removed to focus evaluation on retrieval-dependent reasoning.

\section{Inter-Annotator Agreement Analysis}
\label{app:annotation_agreement}

\begin{figure}
\centering
\includegraphics[width=0.8\linewidth]{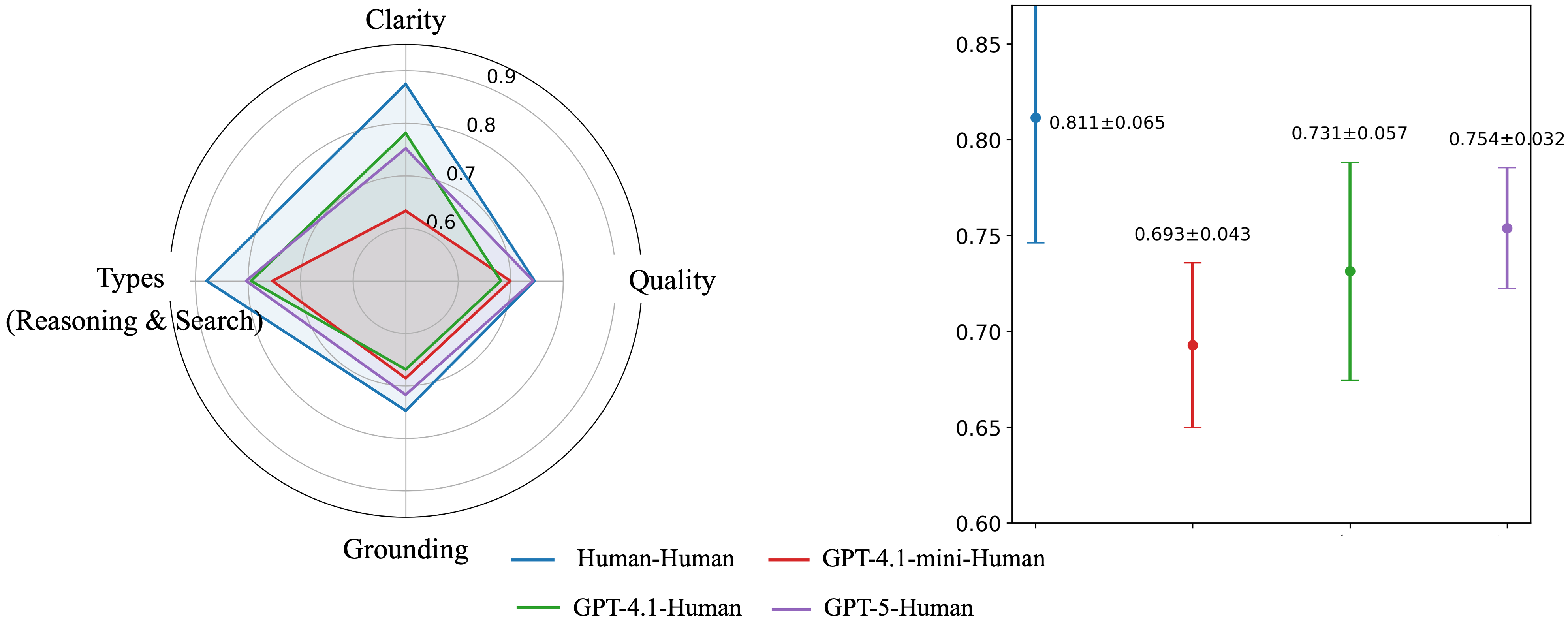}
\caption{Inter-annotator agreement for \algo. (\textit{Left}) Per-field annotation agreement across different competency dimensions. (\textit{Right}) Average Cohen's $\kappa$ comparing human annotators with GPT-4.1, GPT-4.1-mini, and GPT-5, demonstrating strong alignment between expert human judgments and advanced LLM assessments.}
\label{fig:annotation_agreement}
\end{figure}

\paragraph{Per-Field Agreement Analysis.} The left panel of Figure~\ref{fig:annotation_agreement} demonstrates robust inter-annotator agreement across all four annotation fields. The \textbf{Functional Type} field achieves the highest agreement ($\kappa > 0.8$), indicating that annotators can reliably distinguish between different reasoning purposes (e.g., Information Synthesis vs. Plan Formation). The \textbf{Quality Attribute} field shows similarly strong agreement ($\kappa > 0.75$), confirming that evaluative judgments of epistemic soundness are consistently interpretable across annotators. These results establish that our schema captures meaningful, distinguishable patterns in agent reasoning behavior rather than subjective interpretations.

\paragraph{Human-LLM Alignment Assessment.} The right panel reveals substantial alignment between human expert judgments and LLM assessments across all three evaluated models. Human annotators achieve the highest overall agreement ($\kappa = 0.811$), establishing the gold standard for annotation quality. Among LLM judges, GPT-5 demonstrates the strongest alignment with human experts ($\kappa = 0.754$), followed by GPT-4.1-mini ($\kappa = 0.731$) and GPT-4.1 ($\kappa = 0.693$). This progressive improvement across model versions suggests that more advanced language models can better approximate human reasoning patterns in epistemic evaluation tasks. 

\paragraph{Implications for Large-Scale Evaluation.} These agreement results establish the feasibility of deploying our annotation framework for comprehensive agent evaluation. The strong human-LLM alignment enables cost-effective scaling of our evaluation methodology, while the robust per-field agreement ensures that competency assessments reflect genuine behavioral differences rather than annotation artifacts. This validation is particularly crucial for our three core competencies (Groundedness, Recovery, and Calibration), as it confirms that these constructs can be reliably measured across diverse reasoning traces and evaluators.

\section{LLM-as-Judge for \algo}
\label{app:annotation_schema}

This section presents the comprehensive LLM-as-judge used in CompetenceBench to evaluate agent reasoning, search behavior, and answer quality. The schema is organized into: reasoning types with grounding evaluation, search behavior, search result quality.

\subsection{Reasoning Type Annotation and Grounding Evaluation}
\label{app:reasoning_annotation}

The reasoning annotation schema categorizes agent reasoning steps into four functional types and evaluates their grounding. This comprehensive evaluation helps identify both the cognitive function of reasoning steps and whether they are properly supported by evidence.

\begin{tcolorbox}[colback=gray!10!white, colframe=gray!50!black,title=Reasoning Type Classification and Grouding Evaluation,
    colback=blue!5!white,
    colframe=blue!75!black,
    fonttitle=\bfseries,
  breakable]
You are an expert cognitive scientist and evidence-based critical thinking expert. Your task is to classify the reasoning type of an agent's step and evaluate its grounding based *only* on the evidence it had at the time.

Context: The Agent's Goal (Original Question):
\begin{verbatim}
{question}
\end{verbatim}

Evidence: The Search Results the Agent Had Access To:
\begin{verbatim}
{search_evidence_json}
\end{verbatim}

Agent's Reasoning Text to Analyze:
\begin{verbatim}
"{reasoning_text}"
\end{verbatim}

\noindent\rule{\textwidth}{0.5pt}

Task:
\begin{enumerate}
\item Classify the reasoning type:
   \begin{itemize}
   \item \textbf{StateAssessment}: Assess the current knowledge state, usually identifying a knowledge gap.
   \item \textbf{PlanFormation}: The agent is forming a plan of action.
   \item \textbf{InformationSynthesis}: Synthesize new information (from search results) to form a conclusion.
   \item \textbf{CritiqueAndCorrection}: Critique existing information or conclusions and propose a correction.
   \end{itemize}

\item Evaluate grounding:
   \begin{itemize}
   \item Extract the atomic factual premises from the step (skip meta/plan-only wording that contains no factual claim).
   \item For each premise, find a direct supporting span in the provided evidence. If no exact or near-verbatim support exists, mark that premise as unmatched.
   \item Decide the label with STRICT rules:
     \begin{itemize}
     \item \textbf{Grounded}: ALL atomic premises are supported by explicit evidence spans.
     \item \textbf{Not Grounded}: ANY atomic premise lacks a supporting span; OR the step contains only meta/plan text without factual premises.
     \end{itemize}
   \end{itemize}
\end{enumerate}

Additional rules for grounding:
\begin{itemize}
\item QUESTION anchor alone is NOT sufficient for Grounded; do not label as grounded solely for restating the task/intent.
\item Superlatives/temporal/quantitative claims (e.g., last/first/only, years, counts) require explicit evidence spans.
\end{itemize}

\noindent\rule{\textwidth}{0.5pt}

Your Final Output:
\begin{verbatim}
{
  "reasoning_type": "StateAssessment",
  "grounding": "Grounded",
  "anchor_type": "EVIDENCE",
  "justification": "brief explanation"
}
\end{verbatim}
\end{tcolorbox}

\subsection{Search Behavior Annotation}
\label{app:search_annotation}

The search annotation schema categorizes agent search queries into four behavioral types:

\begin{tcolorbox}[colback=gray!10!white, colframe=gray!50!black,title=Search Behavior Classification,
    colback=green!5!white,
    colframe=green!75!black,
    fonttitle=\bfseries,
  breakable]
You are an expert information retrieval specialist. Your task is to classify the type of search query issued by the agent.

Current Search Query:
\begin{verbatim}
{current_query}
\end{verbatim}

Previous Search Query (if any):
\begin{verbatim}
{previous_query}
\end{verbatim}

\noindent\rule{\textwidth}{0.5pt}

Task:
Classify the search query type:
\begin{itemize}
\item \textbf{InitialQuery}: The agent is issuing its first query in a reasoning chain.
\item \textbf{RefinedQuery}: The agent is refining a previous query based on new information.
\item \textbf{FollowUpQuery}: The agent is asking a follow-up question that is not a direct refinement.
\item \textbf{RepeatQuery}: A query that is the same as the previous query.
\end{itemize}

\noindent\rule{\textwidth}{0.5pt}

Your Final Output:
\begin{verbatim}
{
  "search_type": "InitialQuery",
  "justification": "brief explanation"
}
\end{verbatim}
\end{tcolorbox}

\subsection{Search Result Quality Assessment}
\label{app:search_result_quality}

This prompt evaluates the quality and clarity of search results retrieved by the agent. It helps identify when agents work with insufficient or ambiguous information.

\begin{tcolorbox}[colback=gray!10!white, colframe=gray!50!black,title=Search Result Analysis Prompt,
    colback=orange!5!white,
    colframe=orange!75!black,
    fonttitle=\bfseries,
  breakable]
You are an expert data analyst. Your task is to evaluate the quality of a search result based on the query that produced it.

Search Query:
\begin{verbatim}
{query}
\end{verbatim}

Search Result Documents:
\begin{verbatim}
{documents_json}
\end{verbatim}

\noindent\rule{\textwidth}{0.5pt}

Your Task:
Analyze the search result's sufficiency and clarity.

1. Information Quality: Does the result contain enough information to likely answer the user's implicit question in the query? Choose one:
   \begin{itemize}
   \item Sufficient: The answer seems to be present.
   \item Insufficient: The answer is likely not here.
   \end{itemize}

2. Information Clarity: Is the information clear or does it create confusion? Choose one:
   \begin{itemize}
   \item Clear: The information is straightforward and addresses one subject.
   \item Unclear: The results mention multiple distinct entities that could match the query (e.g., two movies with the same title) or the information is vague.
   \end{itemize}

\noindent\rule{\textwidth}{0.5pt}

Your Final Output:
Your response must be a single, valid JSON object with the following attributes:
\begin{itemize}
\item \texttt{information\_quality}: Either "Sufficient" or "Insufficient"
\item \texttt{information\_clarity}: Either "Clear" or "Unclear"
\item \texttt{clarity\_justification}: Brief explanation for your clarity rating
\end{itemize}
\end{tcolorbox}

\section{Accuracy-level performance}\label{app:accuracy}
\begin{table}[h]
\centering
\begin{tabular}{l c}
\toprule
\textbf{Agent} & \textbf{Overall F1 (\%)} \\
\midrule
ASearcher     & 39.77 \\
Search-R1    & 39.29 \\
ReSearch      & 38.30 \\
Few-shot        & 36.04 \\
DeepResearcher & 36.00 \\
Base             & 33.5 \\
\bottomrule
\end{tabular}
\caption{Overall F1 performance across agent variants. Trained agents consistently outperform the base model, with ASearcher achieving the highest score.}
\label{tab:overall_f1}
\end{table}

Table~\ref{tab:overall_f1} reports the aggregate F1 scores across all evaluated agents. We observe that all trained agents outperform the base Qwen model, with ASearcher achieving the best performance (39.8\%). Search-R1 and ReSearch follow closely, while Few-shot prompting and DeepResearcher attain comparable scores. 

\section{Evidence-Grounded Reasoning Analysis}\label{app:evidence_grounded_reasoning}

This section provides detailed analysis of evidence-grounded reasoning capabilities for~\Cref{sec:o1_reasoning}, examining how agents ground their reasoning in retrieved evidence and identifying critical gaps in epistemic alignment. We present two complementary analyses: (1) evidence-conditioned reasoning quality across different evidence states, and (2) type-specific reasoning capabilities that reveal heterogeneous grounding patterns across reasoning skills. 

\subsection{Evidence-Aligned Reasoning: Do Agents Ground Their Inference in What They Know?}\label{sec:rqi_alignment}

\begin{figure}
    \centering
    \includegraphics[width=0.9\linewidth]{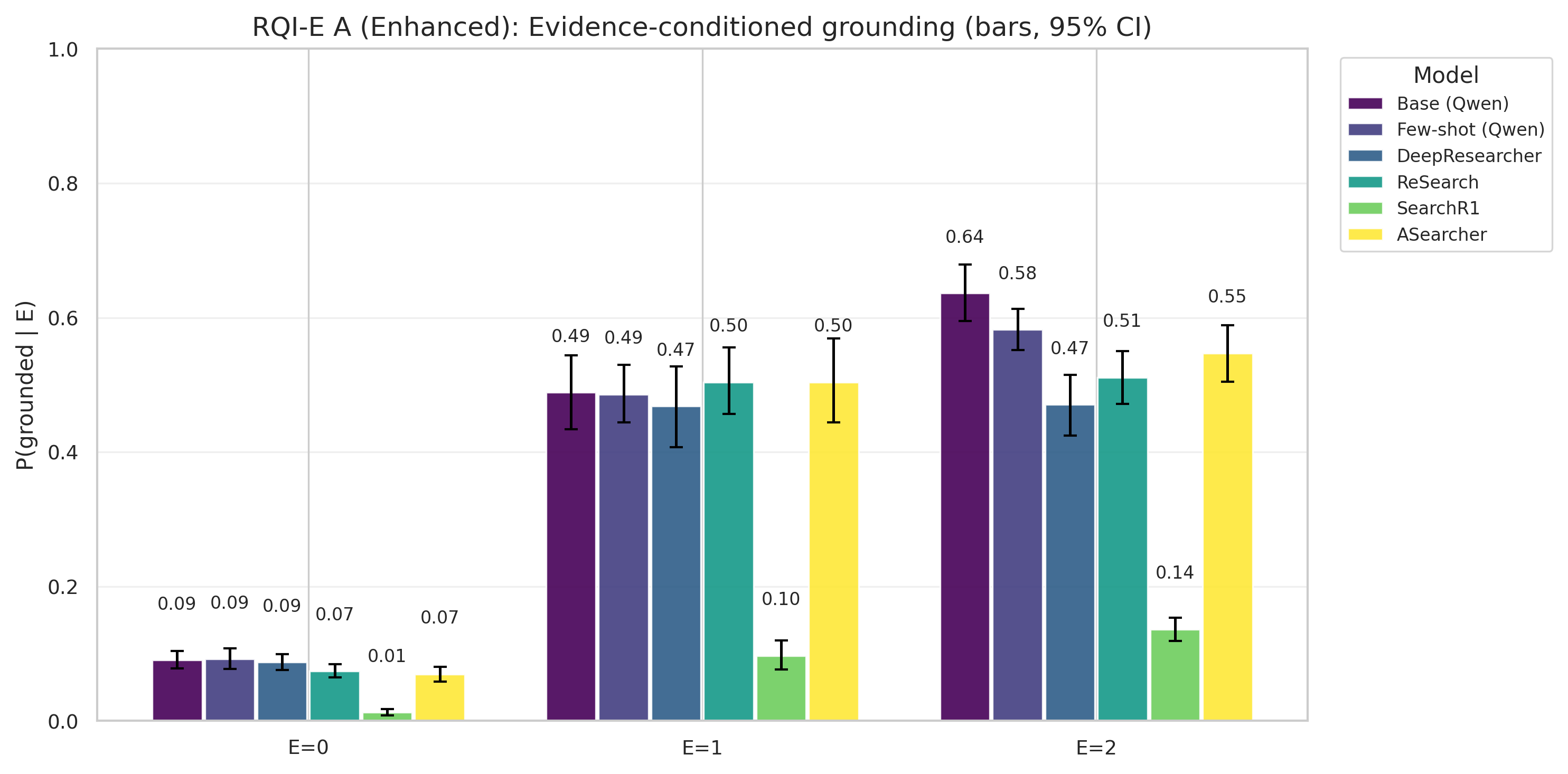}
    \caption{Evidence-conditioned reasoning quality for evidence state $k \in {0,1,2}$ across search agents. Bars denote 95\% confidence intervals. The quantity reflects the expected groundedness of reasoning steps given the epistemic evidence state $E$. Higher values at $E=2$ indicate effective evidence utilization.}
    \label{fig:rqi_model}
\end{figure}
To evaluate whether agents ground their reasoning in retrieved evidence, we analyze the expected groundedness of reasoning steps conditioned on the agent's \textbf{evidence state} $E \in \{0,1,2\}$. The quantity $\mathbb{E}[G_{i,t} \mid E_{i,t}=k]$ measures how reliably an agent produces well-supported reasoning at each evidence level $k$. An \textbf{epistemically sound} agent should avoid unsupported reasoning when $E=0$, provide partial grounding at $E=1$, and fully leverage complete evidence when $E=2$. This conditional analysis enables assessment of \textbf{epistemic alignment}: whether agents reason more confidently only when they possess sufficient evidence.

\paragraph{Base models show better epistemic alignment than specialized agents.} 
Empirical results in Figure~\ref{fig:rqi_model} reveal substantial variation across models. Most agents demonstrate appropriate behavior at $E=0$, with $\mathbb{E}[G \mid E=0] \approx 0.07$–$0.09$, indicating minimal hallucinated reasoning. However, \textsc{Search-R1} exhibits significant \textbf{epistemic misalignment}, with elevated groundedness even under insufficient evidence ($\approx 0.10$ at $E=1$ and $0.14$ at $E=2$), suggesting grounded reasoning. In contrast, \textsc{Base} and \textsc{Few-shot} variants demonstrate the clearest \textbf{evidence-conditioned reasoning}, with groundedness rising from $0.49$ ($E=1$) to $0.64$ ($E=2$), indicating effective epistemic modulation. \textsc{ASearcher} also shows notable improvement ($0.50 \rightarrow 0.55$), while \textsc{ReSearch} and \textsc{DeepResearcher} stagnate around $0.47$–$0.51$, failing to capitalize on stronger evidence. These results demonstrate the necessity for evaluation metrics like \textbf{RQI} that isolate whether reasoning reflects the agent's actual knowledge state.

\paragraph{Case Study: Correct Answer with Ungrounded Reasoning.}
We examine a case where a ReSearch agent correctly answers ``Who won the first celebrity big brother on channel 5?'' despite completely ungrounded reasoning.
After retrieved the first evidence, the agent retrieves conclusive evidence stating ``Celebrity Big Brother 1... concluded on 16 March 2001 when comedian Jack Dee was crowned the winner.''
Despite having the answer, the agent \textit{ignores this evidence and conducts unnecessary searches}, stating: ``I need to clarify which Big Brother series I am referring to... Now, I have to find out the winner of that show.'' The agent eventually answers ``Jack Dee'' correctly, but through an epistemically unsound process. This demonstrates why accuracy metrics alone fail to capture critical reasoning deficiencies.

\subsection{Type-Specific Evidence Alignment: Which Reasoning Skills Are Evidence-Grounded?}
\label{sec:type_rqi_analysis}

\begin{figure}
    \centering
    \includegraphics[width=1\linewidth]{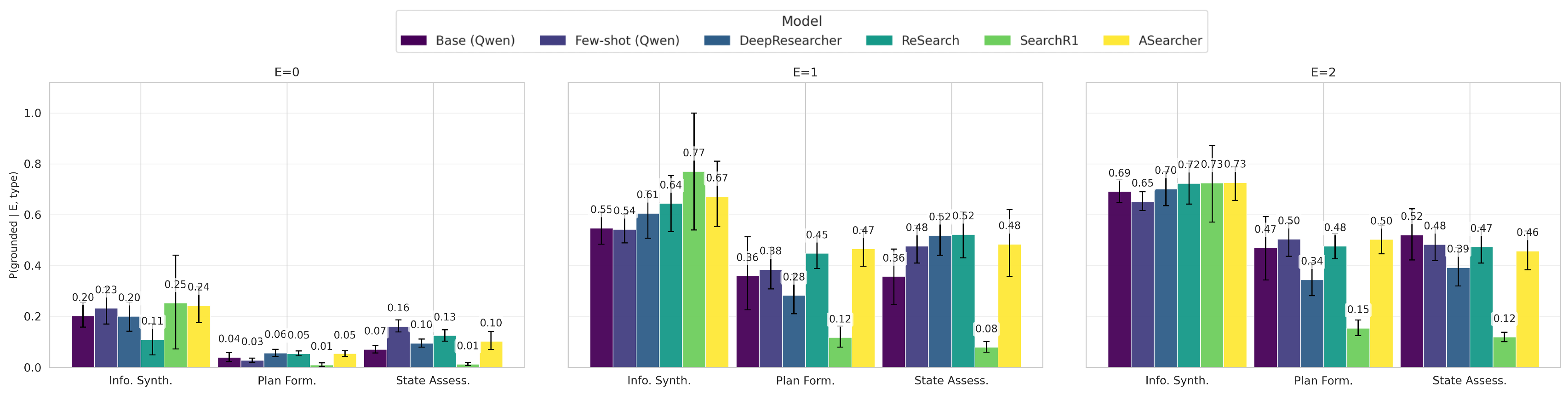}
    \caption{\textbf{Type-Level Evidence-Conditioned Groundedness.} 
Expected groundedness for each reasoning type $\tau$ (Information Synthesis, Plan Formation, State Assessment) under evidence levels $E=0,1,2$. Bars show 95\% confidence intervals. Models exhibit heterogeneous capabilities in grounding specific reasoning skills in available evidence.}
\label{fig:type_rqi_model}
\end{figure}

We further decompose agent reasoning groundedness by reasoning type $\tau \in \{\texttt{IS}, \texttt{PF}, \texttt{SA}\}$, leveraging the \emph{Type-Level Reasoning Quality Index} from Definition~\ref{def:rqi_type} and the evidence-state decomposition in~\Cref{eq:rqi_type_evidence_decomposition}.

Figure~\ref{fig:type_rqi_model} illustrates the groundedness of each reasoning type across three evidence states. Our analysis reveals several significant patterns:

First, \textbf{Information Synthesis (IS)} demonstrates the strongest evidence-responsiveness across all models. With complete evidence ($E{=}2$), IS steps achieve superior groundedness, indicating robust capabilities in aggregating retrieved information. Even with partial evidence ($E{=}1$), agents maintain moderate IS groundedness, suggesting effective utilization of incomplete knowledge.

In contrast, \textbf{Plan Formation (PF)} and \textbf{State Assessment (SA)} exhibit substantially lower \textit{groundedness} even with complete evidence. For PF, only \textsc{ASearcher} and \textsc{ReSearch} exceed 0.5 at $E{=}2$, while others (e.g., \textsc{Search-R1}, \textsc{DeepResearcher}) remain below 0.4, revealing \textit{fragile decision-making processes} despite available knowledge. 
Similarly, \textbf{SA} demonstrates critical limitations: although scores improve under $E{=}2$, most models \textit{underperform} relative to IS, with several agents (e.g., \textsc{Search-R1}, \textsc{DeepResearcher}) showing \textit{minimal evidence-responsiveness} between $E{=}1$ and $E{=}2$. Notably, \textsc{Search-R1} performs comparatively well for IS across all evidence levels but demonstrates \textit{exceptionally poor grounding} for PF and SA (0.15 and 0.12 at $E{=}2$, respectively), suggesting \textit{specialized evidence synthesis} but \textit{narrowly constrained reasoning capabilities}.

These findings demonstrate that only specific reasoning capabilities (particularly synthesis) are consistently \textit{grounded} in retrieved information. This underscores the necessity for developing \textit{evidence-grounded reasoning policies}, especially for higher-order cognitive functions like plan formation and state assessment that currently show significant \textit{epistemic disconnection} from retrieved knowledge.

\section{Evidence Calibration Analysis}
\label{app:calibration_model_breakdown}

To further understand the epistemic calibration capacity of RL-trained agents, we analyze four RL-based agents: \textsc{ASearcher}, \textsc{Search-R1}, \textsc{ReSearch}, and \textsc{DeepResearcher}. Each model exhibits distinct behavior patterns in evidence-grounded answering.

\begin{figure}[ht]
    \centering
    \includegraphics[width=0.95\linewidth]{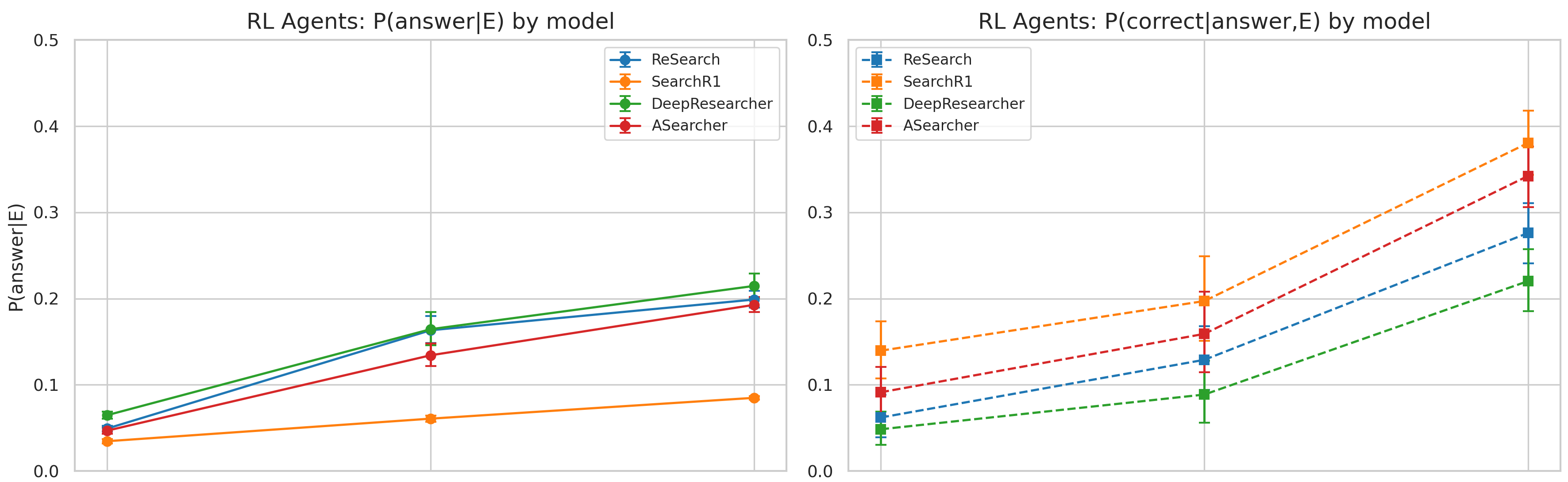}
    \caption{
    \textbf{Evidence-Calibrated Answering by RL Agents.} \textit{Left:} Answering propensity for each RL agent under different evidence levels (E=0/1/2). \textit{Right:} Answering accuracy for each RL agent under different evidence levels (E=0/1/2).}
    \label{fig:trained_model_breakdown}
\end{figure}

As shown in Figure~\ref{fig:trained_model_breakdown} (\textit{Left}), all agents show increased answering with stronger evidence, suggesting basic epistemic alignment, but they vary in \emph{evidence gradient}—the increase from E=0 to E=2. \textsc{ASearcher} and \textsc{DeepResearcher} exhibit higher gradients, indicating \textit{stronger sensitivity to epistemic evidence}. However, all agents maintain relatively low absolute response rates even with sufficient evidence. \textsc{ASearcher} and \textsc{DeepResearcher} reach 19–21\%, while \textsc{Search-R1} remains at 8.5\%, suggesting more conservative behavior.

\paragraph{Calibration vs Accuracy}
As noted in~\Cref{app:accuracy}, \textsc{ASearcher} achieves the highest overall F1 score, followed by \textsc{Search-R1}, \textsc{ReSearch}, and \textsc{DeepResearcher}.

To evaluates whether agents can defer or respond based on the epistemic adequacy of observed evidence. 
As shown in Figure~\ref{fig:trained_model_breakdown}, across trained agents, we observe distinct calibration profiles:

\begin{itemize}[leftmargin=2em,itemsep=0pt,parsep=0pt,topsep=0pt]
\item \textsc{ASearcher} demonstrates superior evidence sensitivity coupled with high accuracy. It responds predominantly when sufficient evidence is available ($E{=}2$) at a substantial rate (19.3\%) while minimizing overconfident responses. As shown in Figure~\ref{fig:trained_model_breakdown} (\textit{Right}), it achieves the second highest conditional accuracy $P(\text{correct}|\text{answer}, E=2)$. This strategic alignment between evidence-based answer timing and correctness yields optimal performance.

\item \textsc{Search-R1} achieves the highest accuracy when answering with complete evidence (Figure~\ref{fig:trained_model_breakdown}, \textit{Right}), but exhibits extreme conservatism, answering in merely 8.5\% of high-evidence states (Figure~\ref{fig:trained_model_breakdown}, \textit{Left}).
While this demonstrates exceptional calibration awareness, the excessive caution significantly constrains overall performance, representing a clear trade-off between coverage and precision.
\end{itemize}

We conclude that well-calibrated agent behavior requires satisfying two critical conditions: (1) deferring responses until evidence strength reaches sufficient levels (e.g., $E{=}2$), and (2) producing correct answers when responding, demonstrating effective utilization of the available evidence. This dual requirement highlights the challenge of balancing epistemic caution with informational utility.

\paragraph{RL Training Reduces Calibration Errors.} Table~\ref{tab:calibration_error_rates_app} provides a detailed breakdown of calibration performance across individual RL-trained agents. While all RL agents show substantial improvements over base models in reducing overconfident answering, \textsc{ASearcher} achieves the lowest overall calibration error (0.302), closely followed by \textsc{DeepResearcher} and \textsc{ReSearch} (both 0.305). Notably, \textsc{Search-R1} exhibits the most conservative behavior with the \textit{lowest} overconfident answeringrate (0.226) but \textit{highest} overcautious rate (0.187), suggesting a trade-off between different types of calibration failures. These results demonstrate that RL training consistently improves evidence-based decision making, though specific training approaches yield different calibration profiles.

\begin{table}[t]
  \centering
  \caption{
  \textbf{Calibration Error Breakdown by Agent Type.}  
  Trajectories categorized as: overconfident answering (answering before sufficient evidence), overcautious abstention (failing to answer despite strong evidence), and overall calibration error. Lower values indicate better calibration. ASearcher show the lowest calibration errors. \textbf{Bold} indicates best performance in each category.\looseness=-1
  }
  \label{tab:calibration_error_rates_app}
  \small
  \begin{tabular}{lccc}
  \toprule
  \textbf{Model} & \textbf{(1) Overconfident} $\downarrow$ & \textbf{(2) Overcautious} $\downarrow$ & \textbf{(3) Calibration Error} $\downarrow$ \\
  \midrule
  Base      & 0.631 & {0.030} & {0.329} \\
  Few-shot  & 0.511 & \textbf{0.024} & 0.317 \\
  \midrule
  ASearcher   &{0.343} & 0.044 & \textbf{0.302} \\
  DeepResearcher & 0.461 & 0.048 & 0.309 \\
  ReSearch & 0.406 & 0.047 & 0.305 \\
  Search-R1 & \textbf{0.226} & 0.187 & 0.319 \\
  \bottomrule
  \end{tabular}
  \end{table}

\begin{figure}[t]
  \centering
  \includegraphics[width=\linewidth]{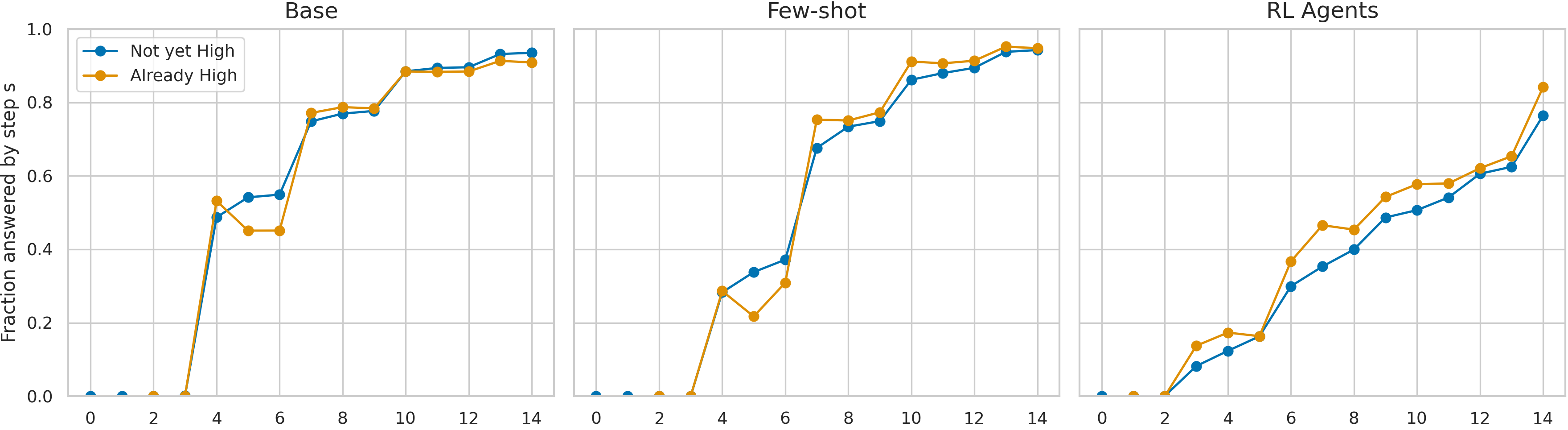}
  \caption{\textbf{Evidence-Conditioned Answer Timing Analysis.}
  For each model group, we plot the fraction of answered trajectories over time (x-axis: turn number), split by whether high evidence ($E{=}2$) has been observed. 
  If agents defer until sufficient evidence, the orange curve (already high) should rise earlier than the blue curve (not yet high). 
  However, all models show little separation between the two, confirming widespread \textit{overconfident answering behavior}.}
  \label{fig:r3e_timing}
\end{figure}

\paragraph{Do agents defer answering until strong evidence arrives?}
To evaluate whether agents appropriately delay answering until they have observed sufficient evidence, we analyze the timing of answers across different evidence states.
Figure~\ref{fig:r3e_timing} presents a temporal analysis comparing the cumulative fraction of answers over time for two distinct trajectory groups:
those where agents have already encountered strong evidence ($E{=}2$) versus those where they have not.

In an \textit{ideally calibrated system}, we would expect agents to predominantly answer after observing strong evidence, resulting in a clear separation between trajectories—specifically, a higher orange curve (evidence already observed) and a lower blue curve (evidence not yet observed).
However, our analysis reveals that \textsc{Base} and \textsc{Few-shot} models demonstrate \textit{minimal separation} between these curves—indicating that answers are generated with similar timing regardless of evidence availability.
RL-trained agents, while showing marginal improvement, still exihibits overconfident answering in 76.5\% of trajectories before reaching sufficient evidence ($E{=}2$).

This finding highlights a \textit{fundamental calibration deficiency}: current models consistently make overconfident decisions without aligning their answer timing with \textit{epistemic sufficiency}.

\section{Agent Synthesis: Leveraging Epistemic Competencies for Answer Generation}
\label{app:agent_synthesis}

\begin{table}[h]
  \centering
  \scriptsize
  \begin{tabular}{l c c c c c c c}
  \toprule
  \textbf{Synthesizer (S)}/ \textbf{Policy (P)} & \textbf{Search-R1} & \textbf{ASearcher} & \textbf{DeepRes.} & \textbf{ReSearch} & \textbf{Base} & \textbf{Fewshot} & \textbf{Overall Avg. $\Delta$F1} \\
  \midrule
  Search-R1 & - & -0.08 & +0.19 & +1.66 & +3.50 & +1.10 & \textbf{+1.27} \\
  ASearcher & +0.38 & - & -0.73 & -0.39 & +0.93 & +1.46 & {+0.33} \\
  DeepRes. & -0.63 & +0.20 & - & +0.00 & +2.24 & +1.45 & {+0.65} \\
  ReSearch & -0.42 & +0.13 & +0.36 & - & +2.99 & +1.89 & {+0.99} \\
  \bottomrule
  \end{tabular}
  \caption{\textbf{Agent Synthesis Performance.} Each cell shows F1 score improvement ($\Delta$F1) when using row agent (S) as synthesizer to generate answers based on evidence collected by column agent (P). Positive values indicate the synthesizer improved upon the original policy's performance. Search-R1 demonstrates the highest overall improvement (+1.06 F1) across all evidence sources, particularly excelling with base (+3.50) and ReSearch (+1.66) evidence collections.}
  \label{tab:synthesizer_results}
  \end{table}

  Our comprehensive evaluation reveals several critical insights into agent capabilities: \textsc{ASearcher} demonstrates superior performance in \textbf{evidence acquisition} and \textbf{recovery mechanisms} (achieving the highest overall F1 score), while \textsc{Search-R1} exhibits exceptional proficiency in \textbf{information synthesis} (with a RQI score of 0.63) coupled with \textbf{minimal overconfident answering behavior} (as discussed in \Cref{sec:exp_calibration}). This observed specialization of epistemic competencies motivated us to explore the potential of \textit{agent synthesis} where we leverage one agent's evidence collection capabilities as input for another agent's answer generation process. 
\paragraph{Synthesizer Evaluation Methodology.}
To test this hypothesis, we evaluate each agent as a \textit{synthesizer} by providing it with the complete reasoning traces, search results, and evidence from other agents' trajectories. The synthesizer's task is to generate a final answer based solely on this information, without performing additional searches. This setup isolates the agent's ability to synthesize information from existing evidence, separate from its search and retrieval capabilities.

\paragraph{Search-R1 Emerges as the Superior Synthesizer.}
As shown in Table~\ref{tab:synthesizer_results}, \textbf{Search-R1} delivers the largest average F1 gain (+1.27), significantly outperforming other agents: ASearcher (+0.33), DeepResearcher (+0.65), and ReSearch (+0.99). This result aligns with our earlier findings that Search-R1 exhibits strong information synthesis capabilities (RQI = 0.63 for Information Synthesis steps) and conservative answering behavior (lowest overconfident answering rate).

The superior synthesis performance of Search-R1 can be attributed to its \textbf{specialized reasoning} capabilities. Despite its low overall RQI (0.08), Search-R1 demonstrates particular strength in \textbf{information synthesis} when provided with clear evidence. Its \textbf{conservative answering behavior}, while limiting coverage in standalone scenarios, becomes an advantage in synthesis tasks where it can carefully evaluate and integrate information from multiple sources before providing a final answer.

\paragraph{Hidden Behaviors: How Accuracy-Level Evaluation Obscures Profound Reasoning Capabilities.}
Surprisingly, \textsc{Base} evidence collection achieved the highest F1 gains when paired with other models for answer generation. This reveals that final accuracy metrics can be misleading when used in isolation, as they obscure critical process-level competencies. While Base appears to have the lowest F1 score when evaluated independently (33.5\% as shown in Table~\ref{tab:overall_f1}), when other models use \textsc{Base}'s evidence as input for answer generation, they can produce substantially better answers (up to +3.50 F1 improvement with Search-R1). This hidden behavior highlights how accuracy-level evaluation alone can mischaracterize agent capabilities and overlook valuable epistemic strengths. The \textsc{Base} may be collecting high-quality evidence but struggling with synthesis, a nuance completely missed by traditional evaluation methods that focus solely on final answer accuracy rather than decomposing the reasoning process into its constituent competencies.

\paragraph{Implications for Agent Design.}
These findings demonstrate that our benchmark and evaluation framework enables \textit{modularization of agent-specific epistemic competencies} to create more effective information-seeking systems. This represents a significant advance for process-level evaluation of agents compared with traditional answer-level evaluation, enabling the identification and combination of complementary strengths across different agent architectures.

\section{Discussion: Understanding Metric Trade-offs}\label{sec:metric_tradeoffs}

Our evaluation framework reveals distinct trade-offs between epistemic competencies that traditional accuracy-only evaluation fails to capture. These insights are critical for both interpreting agent performance and designing effective systems.

\paragraph{The Accuracy-Reasoning Trade-off.}
We observe a particularly concerning inverse relationship between answer accuracy (F1) and reasoning quality (RQI) among RL-trained agents. While RL training improves final answer correctness and calibration, it simultaneously degrades evidence-grounded reasoning quality. This reveals a fundamental tension: \emph{agents can be optimized to produce correct answers without developing sound reasoning capabilities}—a critical consideration for AI safety and interpretability.

\paragraph{Calibration vs. Reasoning Quality.}
RL-trained models demonstrate better calibration (lower CE) despite worse reasoning groundedness (lower RQI), indicating that \emph{well-calibrated agents may still produce poorly justified reasoning}. This highlights the necessity of evaluating both when to answer (calibration) and how to reason (groundedness) as separate competencies.

\paragraph{Implications for Agent Selection and Design.}
These trade-offs directly impact deployment decisions: applications requiring high accuracy may favor \textsc{ASearcher} despite reasoning limitations; those requiring interpretable reasoning may prefer Few-shot models despite lower accuracy; and applications demanding both may benefit from agent synthesis approaches (Section~\ref{sec:exp_agent_usage}). Future agent development should explicitly address these trade-offs through multi-objective optimization and potentially modular architectures that separate evidence acquisition, reasoning, and decision-making components rather than optimizing solely for accuracy.

\end{document}